\documentclass[final,5p,times,twocolumn]{elsarticle}
\usepackage{amssymb}
\usepackage{amsmath}
\usepackage{url}
\usepackage{colortbl}   
\usepackage{enumitem}
\usepackage{amsmath,amssymb}
\usepackage{geometry}
\usepackage{pifont}
\usepackage{graphicx}
\usepackage{booktabs}

\usepackage{caption}
\usepackage{cleveref}
 
\Crefname{figure}{Fig}{Figures}

\makeatletter
\def\ps@pprintTitle{%
  \let\@oddhead\@empty
  \let\@evenhead\@empty
  \def\@oddfoot{\reset@font\hfil\thepage\hfil}%
  \let\@evenfoot\@oddfoot
}
\makeatother

\begin{document}

\begin{frontmatter}



\title{Generative 3D Gaussian Splatting \\ for Arbitrary-ResolutionAtmospheric Downscaling and Forecasting}

%% 作者列表
\author[label1,label2]{Tao Han\fnref{fn1}}
\author[label3]{Zhibin Wen\fnref{fn1}}
\author[label4]{Zhenghao Chen}
\author[label2]{Fenghua Lin}
\author[label5]{Junyu Gao} 
\author[label1]{Song Guo\corref{cor1}}
\ead{songguo@ust.hk} 
\author[label2]{Lei Bai\corref{cor1}}
\ead{bailei@pjlab.org.cn}

\fntext[fn1]{These authors contributed equally to this work and share first authorship.}
\cortext[cor1]{Corresponding authors.}

\affiliation[label1]{organization={Department of Computer Science and Engineering, The Hong Kong University of Science and Technology},
            % addressline={Clear Water Bay, Kowloon},
            % city={},
            postcode={999077},
            state={Hong Kong},
            country={China}}

\affiliation[label2]{organization={Shanghai Artificial Intelligence Laboratory},
            city={Shanghai},
            postcode={200232},
            country={China}}

\affiliation[label3]{organization={Department of Computer Science and Engineering, Southern University of Science and Technology},
            % addressline={1088 Xueyuan Avenue, Nanshan District},
            city={Shenzhen},
            postcode={518055},
            state={Guangdong},
            country={China}}

\affiliation[label4]{organization={School of Computer and Information Sciences, University of Newcastle},
            % addressline={Callaghan},
            city={Newcastle},
            postcode={2308},
            state={NSW},
            country={Australia}}

\affiliation[label5]{organization={School of Artificial Intelligence, OPtics and ElectroNics (iOPEN), Northwestern Polytechnical University},
            % addressline={127 West Youyi Road, Beilin District}, 
            city={Xi'an},
            postcode={710072}, 
            state={Shaanxi},
            country={China}}

%% Abstract
\begin{abstract}
While AI-based numerical weather prediction (NWP) enables rapid forecasting, generating high-resolution outputs remains computationally demanding due to limited multi-scale adaptability and inefficient data representations. We propose the 3D Gaussian splatting-based scale-aware vision transformer (GSSA-ViT), a novel framework for arbitrary-resolution forecasting and flexible downscaling of high-dimensional atmospheric fields. Specifically, latitude–longitude grid points are treated as centers of 3D Gaussians. A generative 3D Gaussian prediction scheme is introduced to estimate key parameters, including covariance, attributes, and opacity, for unseen samples, improving generalization and mitigating overfitting. In addition, a scale-aware attention module is designed to capture cross-scale dependencies, enabling the model to effectively integrate information across varying downscaling ratios and support continuous resolution adaptation. To our knowledge, this is the first NWP approach that combines generative 3D Gaussian modeling with scale-aware attention for unified multi-scale prediction. Experiments on ERA5 show that the proposed method accurately forecasts 87 atmospheric variables at arbitrary resolutions, while evaluations on ERA5 and CMIP6 demonstrate its superior performance in downscaling tasks. The proposed framework provides an efficient and scalable solution for high-resolution, multi-scale atmospheric prediction and downscaling. Code is available at: \url{https://github.com/binbin2xs/weather-GS}. 
\end{abstract}

%% Keywords

\end{frontmatter}



\section{Introduction}

\begin{figure*}[t]
	\centering
	\includegraphics[width=\linewidth]{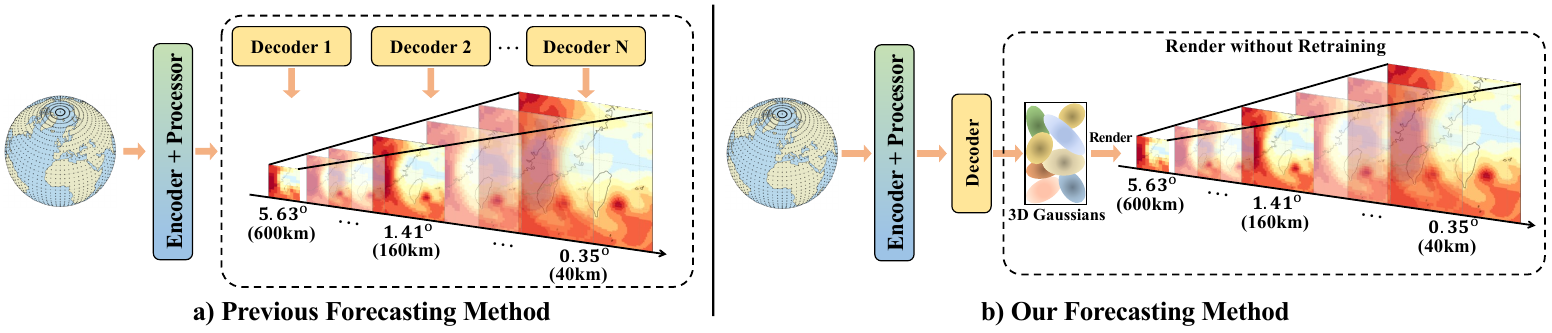}
	\caption{Comparison of arbitrary-resolution atmospheric forecasting methods. (a) Previous methods require separate decoders for each resolution (e.g., 600 km, 160 km, 40 km), increasing model parameters and GPU usage. (b) Our method predicts the continuous 3D Gaussians using a single decoder. Arbitrary resolutions can then be rendered directly without retraining, enabling efficient arbitrary-resolution forecasting while reducing model complexity and GPU memory cost.}
        \label{fig:intro}
\end{figure*}

Atmospheric downscaling and weather forecasting are cornerstone tasks in modern atmospheric science, supporting economic activities, public safety, and disaster preparedness~\cite{vandal2017deepsd, mardani2025residual, Lam2023, Bi2023}. Despite recent progress, most existing downscaling methods are constrained by fixed-scale training paradigms, limiting their applicability when target resolutions vary across regions or tasks. Artificial Intelligence (AI)-based Numerical Weather Prediction (NWP) models still face critical limitations~\cite{Lam2023, Bi2023, Chen2023, Kochkov2023,chen2025va,chen2025stcast}. Most are built for fixed spatial resolutions (e.g., 0.25°) and lack the flexibility to adapt across scales, restricting their effectiveness in tasks ranging from localized storm tracking to global climate modeling \cite{Brenowitz2019}. This limitation is especially concerning in light of the increasing frequency of extreme weather events, such as hurricanes, heatwaves, and heavy rainfall, driven by climate change, which necessitates high-resolution, multi-scale forecasting. In practice, extending current downscaling and forecasting systems to support arbitrary resolutions is computationally expensive. High-resolution modeling inherits the cost of solving large-scale partial differential equations, and resolution flexibility typically requires training separate models or resolution-specific decoders for each target grid. As shown in Fig.~\ref{fig:intro}, supporting a wider range of super-resolution targets leads to rapidly growing model size and GPU memory consumption, resulting in poor scalability, high computational cost, and pressing a need for efficient climate data representations and compression~\cite{han2025climate}.

% To address this issue, we propose a novel approach, GaussianCast that employ 3D Gaussian Splatting (3DGS) techinque for arbitrary-scale numerical weather forecasting, leveraging its continuity, flexibility, and computational efficiency inherits from the real-time radiance field rendering study \cite{Kerbl2023}. 
% %
% In order to obtain a optimal 3DGS location and avoid excessive point density near the poles, leading to computational redundancy, we adopt the the classical Reduced Gaussian Grid (RGG) strategy used by European Centre for Medium-Range Weather Forecasts (ECMWF)~\cite{Bauer2015}\footnote{ECMWF applies RGG to its Integrated Forecast System (IFS) from regular latitude-longitude grids to more efficient designs.} to reduce the number of grid points at high latitudes while preserving physical consistency. 
% %
% By modeling RGG grid points as Gaussian centers, 3DGS enables a compact and continuous representation of atmospheric fields, reducing high-latitude redundancy while preserving physical fidelity. The continuous nature of Gaussian distributions allows for seamless resampling at any resolution, supporting multi-scale forecasting from regional to global scales, from hundreds of kilometers to several kilometers, without polar distortion. 

To address these limitations, we propose a novel framework for arbitrary-resolution atmospheric downscaling and forecasting based on 3D Gaussian Splatting (3DGS). Our method leverages the continuity, flexibility, and computational efficiency of 3DGS, as demonstrated in recent real-time radiance field rendering studies \cite{Kerbl2023}.
To apply 3D Gaussian Splatting to atmospheric reconstruction for downscaling and forecasting, the placement of 3D Gaussians must be defined in a spatially consistent and computationally efficient manner. In this work, we adopt a latitude-longitude grid and align the centers of the 3D Gaussians with the grid points. With the Gaussian centers fixed, atmospheric fields are represented through the key 3DGS parameters, including covariance matrices, attributes, and opacity. This yields a compact and continuous representation of atmospheric data while preserving physical fidelity. The inherent continuity of Gaussian distributions enables seamless resampling at arbitrary resolutions, supporting multi-scale downscaling and forecasting from regional to global levels, spanning spatial resolutions from kilometers to hundreds of kilometers without requiring resolution-specific retraining.

% With optimized 3DGS location, we can perform weather forcasting by generateing new 3DGS key parameters. However, most previous 3DGS methods requires overfitting individual samples and lacking generalization capability for generating unseen samples~\cite{Zhang2024GaussianImage, Zhang2024ImageGS}. This prevents the accurate weather prediction. Insipred by some methods~\cite{huang2024gaussianformer, dong2025gaussiantoken}, we introduce a generative 3D Gaussian prediction method for generating 3DGS parameters. 
% %
% Specifically, we use employs a generative 3D Gaussian prediction approach, which uses Multi-scale Graph Attention Transformers (GATs) to dynamically produce key 3DGS parameters (covariance matrices, attributes, and occupancy) based on XXXX. This network effectively capture the chaotic nature of weather systems with varying spatial influences
%

With optimized 3DGS locations, atmospheric downscaling and forecasting can be performed by generating new 3DGS key parameters. However, most existing 3DGS methods rely on overfitting individual samples and lack the generalization capability needed to produce unseen instances \cite{Zhang2024GaussianImage, Zhang2024ImageGS}, which limits their effectiveness in accurate atmospheric downscaling and weather prediction. Inspired by recent advances in generative 3DGS methods \cite{huang2024gaussianformer}, we propose a generative 3D Gaussian framework to synthesize 3DGS parameters for new samples.
Specifically, our approach employs 3DGS-based Scale-Aware Vision Transformer (GSSA-ViT), a ViT augmented with scale-aware cross attention. By injecting the scale embedding into the cross attention module, the model explicitly conditions feature representations on the target resolution, enabling resolution-adaptive modulation for both downscaling and prediction. Conditioned on the latitude-longitude grid location and observed atmospheric variables, GSSA-ViT dynamically generates essential 3DGS parameters, including covariance matrices, attributes, and opacity, supporting robust and flexible multi-scale atmospheric modeling.

% Our approach addresses the scientific challenge of integrating 3DGS's continuous representation with RGG's physical rationality, enabling efficient, interpretable, and flexible weather forecasting. 

We conduct extensive experiments to evaluate GSSA-ViT on the ERA5 reanalysis dataset \cite{Hersbach2020} and CMIP6 simulations \cite{eyring2016overview}. The results demonstrate that GSSA-ViT significantly reduces arbitrary-scale reconstruction errors while providing a compact and efficient representation of high-dimensional atmospheric data. Importantly, our method supports multi-scale supervision during training by generating Gaussian parameters at arbitrary resolutions, enabling resolution-adaptive learning. In contrast, existing forecasting models are trained at fixed resolutions and can only produce higher-resolution outputs via interpolation. In medium-range forecasting, GSSA-ViT achieves arbitrary-resolution predictions that surpass the performance of such interpolated models, providing more accurate prediction with lower computational cost.

% Our approach addresses the scientific challenge of combining 3DGS's continuous representation with RGG's physical rationality to achieve efficient, interpretable, and flexible weather forecasting. Specifically, we model each RGG grid point as the center of a 3D Gaussian distribution, with parameters (position, covariance, intensity) optimized to represent 160 atmospheric variables, such as temperature, humidity, and wind speed. The training process utilizes historical weather data, such as the ERA5 reanalysis dataset \cite{Hersbach2020}, to minimize reconstruction errors from regular grids to RGG, enabling a compressed representation of atmospheric fields. This representation supports multi-scale predictions by adjusting the scale of Gaussian distributions, allowing for focused local forecasts or broad global overviews.

\begin{itemize}[leftmargin=0pt, itemindent=*]
    \item  We introduce a novel framework that models atmospheric data using 3D Gaussian splatting (3DGS), leveraging their continuity to enable arbitrary-resolution atmospheric downscaling while providing a compact and expressive representation.
    \item  We propose the Gaussian distribution-based weather forecasting paradigm, transforming 3DGS from fitting to prediction, enhancing its generalization and enabling the 3DGS for forecasting tasks.
    \item  We achieve improved performance on atmospheric downscaling tasks and develop the first medium-range forecasting model capable of arbitrary-resolution predictions, achieving competitive results compared to fixed-resolution models upsampled via interpolation, highlighting the potential of our paradigm as a new research frontier.
\end{itemize}

\section{Related Work}
% \paragraph{Numerical Weather Prediction}
% Traditional NWP relies on data assimilation, which integrates observational data into initial conditions, followed by solving partial differential equations (PDEs) to simulate atmospheric dynamics \cite{Lynch2008}. The computational cost scales exponentially with resolution, prompting the adoption of optimized grid systems. The Reduced Gaussian Grid (RGG) addresses high-latitude redundancy by reducing grid points near the poles, as implemented in ECMWF's T1279 model, achieving approximately 40\% computational savings \cite{Hortal1991}. Despite its efficiency, RGG's integration with AI remains underexplored.
\textbf{Atmospheric Downscaling.}
Climate downscaling aims to derive high-resolution climate information from coarse-resolution global climate model outputs. Early approaches primarily relied on dynamical and statistical techniques \cite{zhang2020comparison}. Dynamical downscaling employs regional climate models nested within global climate models and driven by their boundary conditions to resolve fine-scale atmospheric processes \cite{xu2019dynamical}, whereas statistical methods establish empirical relationships between large-scale predictors and local climate variables \cite{wilby1998statistical}. Despite their success, these approaches face several limitations, including high computational cost for dynamical models and strong stationarity assumptions in statistical methods. 

These limitations have motivated the exploration of deep learning approaches for climate downscaling \cite{hohlein2020comparative, liu2020climate, wang2021deep}. Neural networks, including convolutional networks, generative adversarial networks, and graph-based architectures, can learn complex, nonlinear mappings from coarse-resolution climate fields to high-resolution local climate fields \cite{vandal2017deepsd, leinonen2020stochastic, blasone2025graph}. Compared with traditional statistical downscaling methods, deep learning models are typically more computationally efficient and less constrained by stationarity assumptions, while effectively capturing intricate spatial patterns and extreme events. However, most existing deep learning methods are designed for fixed downscaling ratios and often focus on a single climate variable, which limits their flexibility and applicability to multi-variable atmospheric fields and arbitrary-resolution predictions. Although some approaches have been proposed for arbitrary-resolution climate downscaling, such as MINet \cite{chen2025arbitrary} and SGD \cite{tu2025satellite}, these methods still have limitations. MINet constructs high-resolution features primarily from local neighborhoods through a multi-scale coordinate retrieval block, which restricts its ability to capture long-range spatial dependencies and global climate patterns. The SGD model heavily relies on external satellite observation data to guide the diffusion process, making it sensitive to data availability. 

In computer vision, super-resolution methods such as FSRCNN \cite{dong2016accelerating} and ESPCN \cite{shi2016real} use de-convolution or pixel-shuffle layers for fast inference but are limited to fixed upscaling factors. Some approaches \cite{zhu2025multi, wang2026taylor} extend super-resolution to arbitrary resolutions, including Meta-SR \cite{hu2019meta}, which predicts high-resolution details at any scale, and LIIF \cite{chen2021learning}, which uses implicit neural representations to map pixel coordinates to RGB values. Similarly, GSASR \cite{chen2025generalized} leverages 2D Gaussians for super-resolution. However, transferring these methods to climate downscaling is challenging due to the complex physical structures and spatiotemporal dependencies of atmospheric data.

Unlike these approaches, our framework explicitly models global atmospheric features without relying on external auxiliary data, enabling flexible arbitrary-resolution generation and extending its applicability beyond downscaling to arbitrary-resolution forecasting across multiple climate variables.

\textbf{AI-Based Weather Forecasting.}
Recent advancements in AI-based weather forecasting have significantly enhanced medium-range prediction capabilities. Early efforts include FourCastNet \cite{kurth2023fourcastnet}, which introduced adaptive Fourier neural operators for global high-resolution forecasts up to 7 days. Subsequently, Pangu-Weather \cite{Bi2023} employed 3D convolutional networks for fast, accurate forecasts from 1 hour to 7 days, while GraphCast \cite{Lam2023} utilized graph neural networks to model spatial correlations, achieving skillful medium-range forecasts up to 10 days, outperforming ECMWF’s High-Resolution Forecast (HRES) on over 90\% of verification targets. Fengwu \cite{Chen2023} extended global medium-range forecasts beyond 10 days, showcasing machine learning’s potential for extended predictions. NeuralGCM \cite{Kochkov2023} introduced a neural general circulation model for medium-range forecasting, followed by GenCast \cite{Price2024}, which enhanced predictions with diffusion-based ensemble forecasting and uncertainty quantification. FengWu-4DVar \cite{Zhang2023} and FengWu-Adas \cite{Chen2023a} integrated data assimilation techniques to explore end-to-end medium-range weather forecasting. Fengwu-GHR \cite{Chen2024} achieves $0.1^{\circ}$ kilometer-scale medium-range predictions with limited high-resolution data, and ExtremeCast \cite{Chen2024a} targets extreme weather events within 7 days. WeatherGFT \cite{xu2024generalizing} combines a PDE kernel and neural networks to generalize weather forecasts to finer temporal scales beyond the training dataset.
Aurora \cite{bodnar2025foundation} integrated multi-source data for enhanced accuracy. Finally, AIFS \cite{Lang2024} and AIFS-CRPS \cite{Lang2024b} from ECMWF combined AI with traditional NWP strengths for medium-range forecasting. 
%ORCA \cite{guo2024data} provided robust medium-range forecasts, and MetMamba \cite{Zhao2024} introduced novel architectures for weather prediction. %
% Prithvi WxC~\cite{schmude2024prithvi} supports diverse weather and climate tasks like forecasting and downscaling.

In general, these models rely on fixed-resolution latitude-longitude grids, limiting their multi-scale adaptability \cite{Brenowitz2019, Reichstein2019}. In contrast, our proposed method leverages 3D Gaussian Splatting (3DGS) for continuous multi-scale representation and efficient computation, addressing these limitations and providing a more flexible and interpretable framework for medium-range weather forecasting at arbitrary resolutions.

\textbf{3D Gaussian Splatting.}
3D Gaussian Splatting (3DGS), introduced for real-time radiance field rendering, represents point clouds as 3D Gaussian distributions parameterized by position, covariance, and opacity \cite{Kerbl2023}. Its adaptive density control and differentiable rasterization enable efficient, high-quality rendering, surpassing Neural Radiance Fields (NeRFs) in speed and scalability for 3D scene reconstruction \cite{Kerbl2023, mildenhall2021nerf, Zhou2024, Cheng2024}. 3DGS has been applied to tasks such as dynamic scene tracking and editable scene synthesis, leveraging its explicit Gaussian representations \cite{Luiten2024, Huang2024}. Recent extensions to 2D Gaussian Splatting have explored image representation and compression, where Gaussian distributions model pixel data with parameters like position, rotation, and scaling \cite{Zhang2024GaussianImage, Zhang2024ImageGS}. For instance, GaussianImage achieves high-fidelity image reconstruction at 1000 FPS, demonstrating the efficiency of Gaussian-based modeling for 2D data \cite{Zhang2024GaussianImage}.

% Despite these advances, Gaussian splatting suffers from limited generalization. Existing 2D Gaussian Splatting methods~\cite{Zhang2024GaussianImage, Zhang2024ImageGS}, are constrained to learning on individual samples, lacking the ability to generalize to new samples for compression or reconstruction. Similarly, while 3DGS excels in scene-specific rendering, it struggles to generalize to unseen scenes \cite{Kerbl2023}. 
% %In NWP, 3DGS’s continuous representation enables arbitrary-scale characterization of high-dimensional atmospheric fields. However, sample-level overfitting would destroy the predictability of the weather. 
% To address this, we propose a generative 3DGS framework that transforms 3DGS to a conditional generation task, enabling generalized 3DGS for rendering multi-scale weather forecasts.

Despite these advances, Gaussian splatting suffers from limited generalization. Existing 2D Gaussian splatting methods~\cite{Zhang2024GaussianImage, Zhang2024ImageGS} are restricted to individual samples and cannot generalize to new inputs for compression or reconstruction. Similarly, while 3DGS performs well in scene-specific rendering, it struggles to generalize to unseen scenes \cite{Kerbl2023}. To address this, we propose a generative 3DGS framework that formulates 3DGS as a conditional generation task, enabling generalized multi-scale weather forecast rendering.

\begin{figure*}[t]
	\centering
	\includegraphics[width=\linewidth]{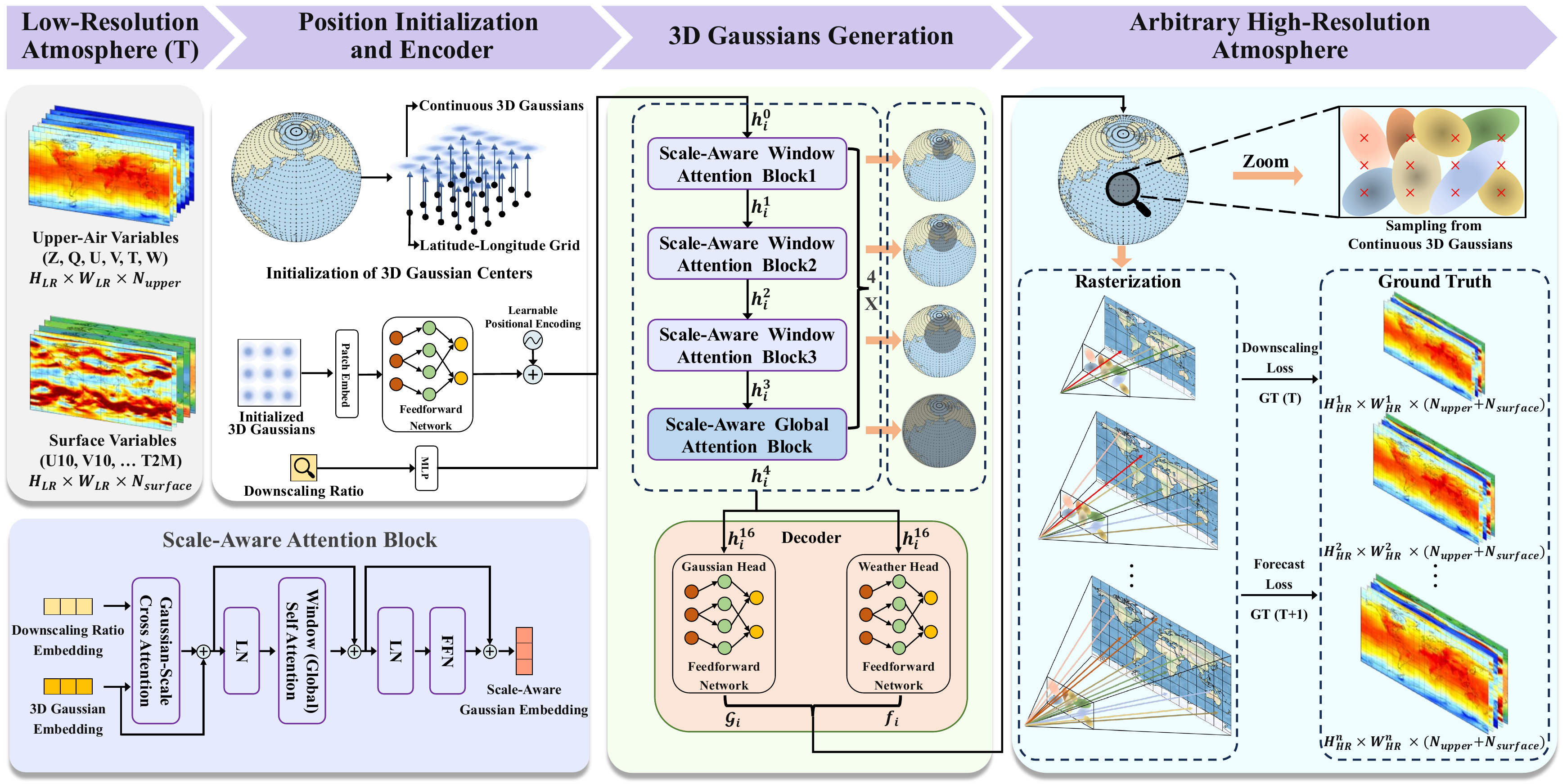}
	\caption{\textbf{Overview of the GSSA-ViT framework.} A low-resolution atmospheric field on a latitude–longitude grid initializes continuous 3D Gaussians, which are encoded as input representations. GSSA-ViT uses scale-aware window attention and global attention to capture resolution-scale information and spatial dependencies, predicting Gaussian parameters. The Gaussians are rendered into high-resolution atmospheric fields at arbitrary resolutions, where different resolutions correspond to different sampling densities.}
        \label{fig:framework}
\end{figure*}

\section{GSSA-ViT: Arbitrary-Resolution Atmospheric Downscaling and Forecasting on Gaussian Space}

% Introducing the novel 3DGS-based method for atmospheric data
% To efficiently represent high-dimensional atmospheric data, we propose a pioneering method that 3DGS \cite{Kerbl2023}, seamlessly integrated with Reduced Gaussian Grids (RGG) \cite{Hortal1991}. This approach marks, to our knowledge, the first application of 3DGS to high-dimensional atmospheric data rendering, delivering a continuous and compact representation of atmospheric fields that supports arbitrary-scale queries and multi-resolution predictions.

% Tailored for the ERA5 dataset's $160 \times 721 \times 1440$ tensor—encompassing 160 atmospheric variables across a 721-latitude by 1440-longitude grid—our method optimizes computational efficiency and adaptability.

\subsection{Atmospheric Data Representation with 3DGS}
\label{sec:3dgs_weatehr}
% Providing context from existing 3DGS literature
\textbf{Basic Concepts of 3DGS.} Originally developed for real-time radiance field rendering, 3DGS represents point clouds as a collection of 3D Gaussian distributions \cite{Kerbl2023}. Each Gaussian is characterized by a position vector $\mu \in \mathbb{R}^3$ defining its center, a covariance matrix $\Sigma \in \mathbb{R}^{3 \times 3}$ determining its shape and orientation, an opacity factor $\alpha \in [0,1)$ for rendering, and spherical harmonics encoding view-dependent color. The method employs adaptive density control to dynamically adjust the number of Gaussians and a fast, differentiable tile-based rasterizer for rendering \cite{Kerbl2023}. Feature 3DGS extends this framework by incorporating high-dimensional semantic feature vectors, enabling tasks like semantic segmentation \cite{Zhou2024}. Inspired by these advancements, we adapt 3DGS to atmospheric data by extending $f$ to an $N$-dimensional feature vector representing $N$ meteorological variables.

% Describing the continuous representation of atmospheric fields
\textbf{Latitude-longitude grid for 3DGS initialization.} We conceptualize the atmospheric field as a function $F: S^2 \to \mathbb{R}^{N}$, where $S^2$ represents the Earth's surface as a unit sphere, and $\mathbb{R}^{N}$ corresponds to $N$ atmospheric variables, such as temperature, humidity, and wind speed. Further details on the atmospheric variables used are provided in the Table~\ref{tab:vars}.
% and we list more details of varibale in Supplementary files
%
% listed in Table~\ref{tab:vars}. 
%
As shown in Fig.~\ref{fig:framework}, we initialize 3D Gaussians on a low-resolution (LR) latitude–longitude grid, consisting of $K$ points $\{p_i\}_{i=1}^K$ defined on the sphere $S^2$. The grid is constructed by uniformly discretizing latitudes $\phi_k \in [-90^\circ, 90^\circ]$ for $k = 1, \dots, H_{\text{LR}}$ and longitudes $\lambda_m \in [-180^\circ, 180^\circ)$ for $m = 1, \dots, W_{\text{LR}}$. Each grid point $p_i$ corresponds to a pair $(\phi_k, \lambda_m)$, forming a regular spherical discretization of the atmospheric field. The atmospheric field is represented by a collection of continuous 3D Gaussians $\mathcal{G} = \{\mathcal{G}_i\}_{i=1}^K$, where each Gaussian $\mathcal{G}_i = (\mu_i, \Sigma_i, f_i, \alpha_i)$ is defined by the probability density function:
\begin{equation}
    \mathcal{G}_i(p) = \alpha_i \cdot \frac{1}{(2\pi)^{3/2} |\Sigma_i|^{1/2}} \exp\left( -\frac{1}{2} (p - \mu_i)^T \Sigma_i^{-1} (p - \mu_i) \right),
\end{equation}
parameterized by its position $\mu_i \in \mathbb{R}^3$, a covariance matrix $\Sigma_i \in \mathbb{R}^{3 \times 3}$, a feature vector $f_i \in \mathbb{R}^{N}$ storing the $N$ variable values at $p_i$, and an opacity factor $\alpha_i \in [0,1)$. The position $\mu_i$ is defined by the corresponding latitude–longitude grid coordinate $(\phi_k, \lambda_m)$ and a fixed vertical coordinate $z_0 = 1$, forming $\mu_i = (\phi_k, \lambda_m, z_0)$, which serves as the center of each Gaussian distribution. The covariance matrix $\Sigma_i$ is constructed as $\Sigma_i = R S S^T R^T$, where $R$ is a rotation matrix parameterized by a quaternion $q_i \in \mathbb{R}^4$, and $S = \text{diag}(s_{i1}, s_{i2}, s_{i3})$ is a diagonal scaling matrix with scaling factors $s_{i1}, s_{i2}, s_{i3}$ along the three axes \cite{Kerbl2023}. This formulation allows the Gaussian to adapt its shape and orientation during optimization. The atmospheric field is thus represented by the collection $\mathcal{G} = \{\mathcal{G}_i\}_{i=1}^K$, enabling a continuous approximation of the data across the sphere.

\subsection{Conditional 3DGS Generation}
\label{subsec:conditional_3dgs}

\textbf{Problem Formulation.} Unlike existing AI forecasting \cite{Lam2023, Bi2023, chen2025arbitrary} and downscaling models, which operate directly on latitude–longitude grids, our framework models the atmospheric state as a collection of Gaussian primitives, enabling a unified formulation for both temporal forecasting and spatial downscaling in Gaussian space. Instead of predicting future Gaussian distributions from the rendered Gaussian space at time $T$, we directly generate the Gaussian distributions at the next time step $T+1$ using the raw atmospheric data at time $T$ as a conditional input. This formulation naturally supports both temporal forecasting and spatial downscaling, differing only in the target time index. Specifically, as depicted in Fig.~\ref{fig:framework}, given the lower-resolution atmospheric field $F_{LR}(T): S^2 \to \mathbb{R}^{N}$ at time $T$, represented as a tensor of shape $N \times H_{\text{LR}} \times W_{\text{LR}}$, our objective is to generate the Gaussian space $\mathcal{G}(T+1) = \{\mathcal{G}_i(T+1)\}_{i=1}^K$, where each $\mathcal{G}_i(T+1) = (\mu_i, \Sigma_i(T+1), f_i(T+1), \alpha_i(T+1))$ is a continuous Gaussian primitive. The generation process is defined as:
\begin{equation} 
\mathcal{G}_i(T+1) = \text{Model}(F_{LR}(T), p_i, \Theta),
\end{equation}
where $p_i$ is the latitude–longitude grid point associated with $\mathcal{G}_i$, $\Theta$ represents the model parameters, and $\text{Model}$ is a learnable neural network consisting of the Gaussian embedding layer, scale-aware attention blocks, and the Gaussian decoding layer.

In the downscaling setting, the formulation remains identical except that the target Gaussian space corresponds to the same time step: $\mathcal{G}(T) = \{\mathcal{G}_i(T)\}_{i=1}^K$, which is generated from a lower-resolution input field $F_{LR}(T)$. Therefore, the only distinction between downscaling and forecasting lies in the temporal index of the target Gaussian distributions: forecasting predicts $\mathcal{G}(T+1)$, whereas downscaling reconstructs $\mathcal{G}(T)$.

\textbf{Gaussian Embedding.} The initial node features are derived from atmospheric data and positional information, using the raw data $F_{LR}(T)$ at time $T$ sampled on latitude--longitude grid points $p_i$. We incorporate a learnable positional embedding $\textbf{E}_{pos} \in \mathbb{R}^{1 \times K \times D}$, where $K$ is the number of grid points and $D$ is the embedding dimension. These embeddings, optimized jointly with the model parameters, are added to the atmospheric feature representations to provide spatial information.

The feature vector $f_i \in \mathbb{R}^{N}$, representing the $N$ atmospheric variables sampled from $F_{LR}(T)$ at $p_i$, is projected into the latent space using a patch embedding layer to produce $h_{\text{f}} \in \mathbb{R}^D$: $h_{f} = \text{PatchEmbed}(f_i).$
The final node feature $h_i^0$ is obtained by adding the learnable positional feature and the atmospheric feature:
\begin{equation} 
    h_i^0 = \textbf{E}_{pos}[i] + h_{f}.
\end{equation}
    
\textbf{Scale-aware Attention Block.} To enable arbitrary-resolution modeling and capture the complex dynamics of the Earth system, we employ a Scale-Aware Attention Block that incorporates resolution information via a downscaling ratio embedding. Given the embedded node features $h^l = \{h^l_i\}^K_{i=1}$, where $h^l_i$ denotes the feature of the $i$-th Gaussian node at layer $l$, the downscaling ratio $r$ is first projected into the latent space using a linear layer to obtain a downscaling ratio embedding $r' \in \mathbb{R}^D: r'=\text{Linear}(r)$. To incorporate resolution information, we apply a cross-attention mechanism where the node features serve as queries and the scale embedding provides the key and value representations: 
\begin{equation} 
    \hat{h}^l=\text{MHA}(h^l,r',r')
\end{equation}
where $\text{MHA}(\cdot)$ denotes multi-head attention. 

To capture both local spatial interactions and long-range dependencies, we employ a combination of window attention and global attention. Window attention models local spatial correlations efficiently, while global attention enables information exchange across distant regions, allowing the model to capture large-scale atmospheric structures. The attention operations update the node features as:
\begin{equation}
    h^{l+1}=\text{Attn}(\hat{h}^l) \in \mathbb{R}^{K \times D}
\end{equation}

By combining local window attention with global attention, the model balances computational efficiency with the ability to capture large-scale spatial dependencies in atmospheric dynamics.

\textbf{Gaussian Decoding.} The updated features $h_i^{L}$ after $L$ layers are decoded using two separate heads to produce the atmospheric variables and the Gaussian parameters of $\mathcal{G}_i(\tau)$. Specifically, we employ two multi-layer perceptron (MLP) heads operating on $h_i^{L}$. The first head predicts the $N$-dimensional feature vector, while the second head outputs the parameters of the Gaussian representation:
\begin{equation}
f_i(\tau) = \text{MLP}_{\text{var}}(h_i^{L}), \qquad
g_i(\tau) = \text{MLP}_{\text{gauss}}(h_i^{L}),
\end{equation}
where $f_i(\tau) \in \mathbb{R}^{N}$ denotes the $N$-dimensional feature vector, and $g_i(\tau) \in \mathbb{R}^{8}$ encapsulates the quaternion for the rotation matrix $R \in \mathbb{R}^{4}$, the scaling factors for the diagonal matrix $S \in \mathbb{R}^{3}$, and the opacity factor $\alpha \in \mathbb{R}^{1}$. These parameters are post-processed: the scaling factors are passed through a softplus activation to enforce positivity, the feature vector and opacity factor through a sigmoid activation to constrain them to physically plausible ranges, and the quaternion is normalized to maintain unit length, reconstructing $\Sigma_i(\tau) = R S S^T R^T$. The position $\mu_i$ remains fixed (as $\mu_i$ is time-invariant per grid point coordinates), so $\mathcal{G}_i(\tau) = (\mu_i, \Sigma_i(\tau), f_i(\tau), \alpha_i(\tau))$. Here, $\tau$ denotes a generic time index. When $\tau = T+1$, the decoded parameters correspond to the forecasted atmospheric state at the next time step. When $\tau = T$, the Gaussian representation is directly used for spatial downscaling of the atmospheric field at arbitrary resolutions.

\subsection{Arbitrary-scale Rendering and Optimization}
\label{subsec:rendering_optimization}

% After generating the Gaussian space $\mathcal{G}(t+1) = \{\mathcal{G}_i(t+1)\}_{i=1}^K$ for the next time step, we perform forecast rendering and optimization to produce predictions at arbitrary scales and ensure accurate reconstruction of atmospheric fields. 

\textbf{Arbitrary-scale Rendering via Rasterization.}
\label{para:forecast_rendering}
As shown in Fig.~\ref{fig:framework}, to render the arbitrary-scale atmospheric field at time $\tau$, we adopt the reconstruction method in Section~\ref{sec:3dgs_weatehr}. Specifically, for any query point $p \in S^2$, the high-resolution atmospheric field $F_{HR}(p, \tau)$ is reconstructed as a weighted sum of feature vectors modulated by opacity:
\begin{equation}
F_{HR}(p, \tau) = \sum_{i \in \mathcal{K}(p)} f_i(\tau) \alpha_i(\tau) w_i,
\end{equation}
where $\mathcal{K}(p)$ is the set of Gaussians overlapping with $p$, sorted by depth, and $w_i = \prod_{j=1}^{i-1} (1 - \alpha_j(\tau))$ is the transmittance ensuring front-to-back accumulation. To support arbitrary-scale predictions, we adjust the resolution of the Gaussian splatting by varying the density and coverage of query points $p$. For high-resolution forecasts (e.g., 0.1$^\circ$ resolution, approximately 10 km), we increase the density of query points to capture fine-grained details, while for lower-resolution forecasts (e.g., 1$^\circ$ resolution, approximately 100 km), we reduce the density, allowing efficient rendering across scales from kilometers to thousands of kilometers. This flexibility leverages the continuous representation of 3DGS and the scale-aware design of the network architecture, enabling GSSA-ViT to seamlessly adapt to diverse spatial scales without retraining.

\textbf{End-to-End Optimization.}
\label{para:end_to_end_optimization}
Given the differentiable nature of 3DGS rendering, we perform end-to-end supervision by directly comparing the rendered forecast $F_{HR}(p, \tau)$ with the ground-truth data $\hat{F}_{HR}(p, \tau)$. The loss function is defined as:
\begin{equation}
\mathcal{L} = \sum_{p \in \text{HRG}} \| F_{HR}(p, \tau) - \hat{F}_{HR}(p, \tau) \|_2^2,
\end{equation}
where $\hat{F}_{HR}(p, \tau)$ represents the high-resolution ground-truth atmospheric field at time $\tau$, with $p$ denoting a spatial coordinate on the high-resolution latitude–longitude grid (HRG). The model parameters $\Theta$ are optimized to generate Gaussian parameters ($\Sigma_i(\tau)$, $f_i(\tau)$, $\alpha_i(\tau)$).

\begin{table*}[t]
  \centering
    \captionsetup{
        format=plain,      
        labelsep=newline,   
        justification=justified, 
        singlelinecheck=false,
        labelfont=bf,
    }
  \caption{A summary of atmospheric variables. Specifically, the upper-air variables are available at 13 standard pressure levels, namely 50, 100, 150, 200, 250, 300, 400, 500, 600, 700, 850, 925, and 1000 hPa.}
  \resizebox{1.0\linewidth}{!}{
    \begin{tabular}{clc|cl|cl}
    \toprule
    \multicolumn{3}{c|}{\textbf{Upper-Air}}   & \multicolumn{4}{c}{\textbf{Surface}} \\
    \midrule
    Name & \multicolumn{1}{c}{Description} & Levels & Name & \multicolumn{1}{c|}{Description}  & Name & \multicolumn{1}{c}{Description}  \\
    \midrule
    
    Z & Geopotential & 13 & U10 & x-direction wind at 10m height & U100 & x-direction wind at 100m height \\
    Q & Specific humidity & 13 & V10 & y-direction wind at 10m height & V100 & y-direction wind at 100m height \\
    U & x-direction wind & 13 & T2M & Temperature at 2m height & TCC & total cloud cover \\
    V & y-direction wind & 13 & MSL & Mean sea-level pressure & D2M & 2-meter dewpoint temperature  \\
    T & Temperature & 13 & TP6H & total precipitation & &\\
    W & Vertical velocity &13 & & & &\\
    
    % \midrule
    % \multicolumn{3}{c}{\textbf{Surface-Incremental Learning}} \\
    % \midrule
    % u10 & x-direction wind at 10m height & Single \\
    % v10 & y-direction wind at 10m height & Single \\
    % t2m & Temperature at 2m height & Single \\
    % msl & Mean sea-level pressure & Single \\
    % sp & Surface pressure & Single \\
    \bottomrule
    \end{tabular}}%
  \label{tab:vars}
\end{table*}

\section{Experiments}
\subsection{Dataset}
\textbf{ERA5.} We use the ERA5 \cite{Hersbach2020} reanalysis dataset produced by the European Centre for Medium-Range Weather Forecasts (ECMWF), which provides global atmospheric fields from 1940 to the present with hourly temporal resolution and 0.25° × 0.25° spatial resolution.

\textbf{CMIP6} We also use climate simulation data from the Coupled Model Intercomparison Project Phase 6 (CMIP6) \cite{eyring2016overview}. Specifically, we use the historical run of the MPI-ESM1-2-LR model, which provides atmospheric variables at 6-hour temporal resolution and a spatial resolution of 1.875° × 1.875°. The historical simulation covers the period from 1850 to 2014 and includes multiple pressure-level variables.

Inputs at resolutions lower than the native resolutions of ERA5 and CMIP6 are generated via bilinear interpolation. For the downscaling task, we consider five commonly used atmospheric variables: geopotential height at 500 hPa (Z500), temperature at 850 hPa (T850), 2 m temperature (T2M), and 10 m wind components (U10, V10). For forecasting, we use six upper-air variables, including geopotential height (Z), specific humidity (Q), zonal wind (U), meridional wind (V), temperature (T), and vertical velocity (W), across 13 pressure levels (50, 100, 150, 200, 250, 300, 400, 500, 600, 700, 850, 925, and 1000 hPa), together with nine surface variables to represent the atmospheric state. The full list of variables is provided in Table~\ref{tab:vars}.

For the CMIP6-to-ERA5 downscaling task, we use 1979–2010 for training, 2011–2012 for validation, and 2013–2014 for testing, with a temporal resolution of 6 hours. The input data for this task is CMIP6 at 5.625° resolution. For ERA5 downscaling, the training, validation, and test periods are 1981–2015, 2016, and 2017–2018, respectively, with a temporal resolution of 1 hour, using ERA5 data at 5.625° resolution as input. For ERA5 arbitrary-resolution forecasting, we train the model on 2000–2019 and evaluate it on 2020–2021, with a temporal resolution of 1 hour, using ERA5 data at 1.40625° resolution as input.

\subsection{Evaluation Metrics.}
Arbitrary-resolution forecast performance is measured using the latitude-weighted root mean square error (LRMSE)~\cite{kurth2023fourcastnet, Chen2024}. For the downscaling task, we report LRMSE, Mean-bias (M-b), and the Pearson coefficient (P). These metrics provide a comprehensive assessment of both the accuracy and reliability.

\textbf{Latitude-Weighted Root Mean Square Error.} The LRMSE addresses the distortion of grid cell areas in latitude-longitude coordinate systems by assigning cosine-latitude weights. For a global field with $N$ grid points, LRMSE is computed as:

\begin{equation}
\text{LRMSE} = \sqrt{ \frac{1}{\sum_{i=1}^{N} w_i} \sum_{i=1}^{N} w_i \cdot (y_i - \hat{y}_i)^2 }
\end{equation}

where $y_i$ and $\hat{y}_i$ are the observed and predicted values at grid point $i$, $w_i = \cos(\phi_i)$ is the weight for grid point $i$, $\phi_i$ is the latitude (in radians) of grid point $i$'s center, and $N$ denotes the total number of grid points.

This weighting balances error contributions across latitudes, as unweighted RMSE would disproportionately emphasize high-latitude grid cells where longitudinal lines converge. The $\cos(\phi)$ weighting exactly compensates for the reduced actual area of grid cells in equal-angle latitude-longitude grids.

\begin{table*}[!t]
\centering
\captionsetup{
    format=plain,           
    labelsep=newline,       
    justification=justified, 
    singlelinecheck=false,
    width=0.89\linewidth,
    labelfont=bf,
}
    \caption{Performance comparison of atmospheric downscaling from MPI-ESM (5.625°) to ERA5 at three target resolutions. The first section reports results at 1.40625°, while the remaining sections evaluate finer resolutions (0.703125° and 0.3515625°). Lower LRMSE indicates better performance, higher P reflects stronger correlation, and M-b values closer to zero indicate smaller bias. The best results are highlighted in \textbf{bold}, and the second-best results are \underline{underlined}.}
\label{tab:downscale1}
\setlength{\tabcolsep}{4pt}

\resizebox{0.89\linewidth}{!}{%
\begin{tabular}{lccc ccc ccc ccc ccc}
\toprule
Methods
& \multicolumn{3}{c}{Z500}
& \multicolumn{3}{c}{T850}
& \multicolumn{3}{c}{T2M}
& \multicolumn{3}{c}{U10}
& \multicolumn{3}{c}{V10} \\

\cmidrule(lr){2-4}
\cmidrule(lr){5-7}
\cmidrule(lr){8-10}
\cmidrule(lr){11-13}
\cmidrule(lr){14-16}

& LRMSE$\downarrow$ & P$\uparrow$ & M-b
& LRMSE$\downarrow$ & P$\uparrow$ & M-b
& LRMSE$\downarrow$ & P$\uparrow$ & M-b
& LRMSE$\downarrow$ & P$\uparrow$ & M-b
& LRMSE$\downarrow$ & P$\uparrow$ & M-b \\
\midrule

\multicolumn{16}{c}{MPI-ESM (5.625$^\circ$) to ERA5 (1.40625$^\circ$)} \\
\midrule

Bicubic
& 1142.43 & 0.92 & 71.36
& 4.80 & 0.93 & 0.11
& 4.07 & 0.97 & \underline{-0.05}
& 5.49 & 0.44 & -0.06
& 5.57 & 0.20 & \textbf{0.00} \\

Bilinear
& 1114.65 & 0.92 & 71.23
& 4.64 & 0.94 & 0.10
& 3.97 & 0.97 & \underline{-0.05}
& 5.24 & 0.45 & -0.06
& 5.34 & 0.20 & \textbf{0.00} \\

\midrule

ResNet \cite{climax, he2016deep}
& 825.75 & \underline{0.96} & -108.54
& 3.60 & \underline{0.96} & 0.19
& 2.89 & \underline{0.98} & 0.14
& 4.05 & 0.65 & 0.06
& 4.11 & 0.45 & 0.09 \\

Unet \cite{climax, ronneberger2015u}
& 858.35 & 0.95 & 35.10
& 3.66 & \underline{0.96} & -0.34
& 2.95 & \underline{0.98} & 0.16
& 4.09 & 0.64 & -0.06
& 4.13 & 0.44 & 0.08 \\

ViT \cite{climax, vit}
& 811.61 & \underline{0.96} & -54.32
& 3.58 & \textbf{0.97} & -0.29
& 2.80 & \textbf{0.99} & -0.06
& 4.01 & \underline{0.66} & -0.08
& 4.07 & \underline{0.47} & \underline{0.01} \\

MetaSR \cite{hu2019meta}
& 791.71 & \underline{0.96} & -11.09
& 3.51 & \textbf{0.97} & \textbf{-0.01}
& 3.06 & \underline{0.98} & \textbf{0.00}
& 3.95 & 0.65 & \textbf{-0.03}
& 3.99 & 0.45 & 0.03 \\

% MetaSR(reproduced(l1loss)) [48]
% & 818.07 & 0.95 & -1.91
% & 3.57 & 0.96 & 0.03
% & 3.20 & 0.98 & 0.28
% & 4.03 & 0.65 & 0.08
% & 4.08 & 0.45 & -0.02 \\

LIIF \cite{chen2021learning}
& 802.60 & \underline{0.96} & 21.30
& 3.50 & \underline{0.96} & \underline{-0.10}
& \underline{2.79} & \textbf{0.99} & 0.14
& 3.92 & \underline{0.66} & 0.13
& 3.98 & 0.46 & -0.07 \\

% LIIF(reproduced(l1loss)) [48]
% & 808.93 & 0.95 & -28.85
% & 3.52 & 0.96 & -0.25
% & 2.87 & 0.98 & -0.38
% & 3.98 & 0.66 & 0.08
% & 4.04 & 0.47 & 0.02 \\

ClimaX \cite{climax}
& 807.43 & \underline{0.96} & \textbf{2.70}
& 3.49 & \textbf{0.97} & -0.11
& \underline{2.79} & \textbf{0.99} & -0.06
& 3.99 & \underline{0.66} & \underline{0.04}
& 4.06 & \underline{0.47} & -0.02 \\

MINet \cite{chen2025arbitrary}
& \underline{786.93} & \underline{0.96} & \underline{-4.67}
& \underline{3.46} & \textbf{0.97} & \underline{-0.10}
& \textbf{2.76} & \textbf{0.99} & -0.18
& \underline{3.87} & \underline{0.66} & 0.07
& \underline{3.94} & \underline{0.47} & \underline{0.01} \\

% MINet (Reproduced(l1loss))
% & 812.49 & 0.95 & -58.05
% & 3.55 & 0.96 & -0.53
% & 2.86 & 0.98 & -0.49
% & 3.98 & 0.66 & 0.05
% & 4.05 & 0.46 & 0.08 \\

% MINet (Reproduced(mseloss))
% & 818.35 & - & -
% & 3.56 & - & -
% & 3.07 & - & -
% & 3.98 & - & -
% & 4.05 & - & - \\

GSASR \cite{chen2025generalized}
& 918.20 & 0.95 & -71.32
& 3.78 & \underline{0.96} & -0.44
& 3.12 & \underline{0.98} & -0.56
& 4.23 & 0.62 & -0.06
& 4.34 & 0.38 & -0.04 \\

GSSA-ViT (Ours)
& \textbf{658.84} & \textbf{0.98} & 85.11
& \textbf{3.20} & \textbf{0.97} & 0.27
& 2.83 & \textbf{0.99} & 0.06
& \textbf{3.71} & \textbf{0.72} & -0.11
& \textbf{3.87} & \textbf{0.56} & 0.02 \\

\midrule
\multicolumn{16}{c}{MPI-ESM (5.625$^\circ$) to ERA5 (0.703125$^\circ$)} \\
\midrule
Bicubic       
& 1141.53 & 0.92  & 71.66 
& 4.80  & 0.93 & 0.11  
& 4.13 & 0.97 & 0.31  
& 5.53 & 0.44 & -0.14 
& 5.58 & 0.20 & \textbf{0.00} \\

Bilinear      
& 1114.30 & 0.92  & 71.53 
& 4.65  & 0.94 & 0.10  
& 4.02 & 0.97 & 0.30  
& 5.28 & 0.45 & -0.14 
& 5.35 & 0.20 & \textbf{0.00} \\

\midrule

ResNet \cite{climax, he2016deep}
& 875.88  & 0.95  & 72.30 
& 3.93  & \underline{0.96} & 0.09  
& 3.84 & 0.97 & 1.08  
& 4.34 & 0.55 & -0.41 
& 4.16 & 0.35 & 0.02 \\

Unet \cite{climax, ronneberger2015u} 
& 980.46  & 0.94  & 83.06 
& 4.11  & 0.95 & -0.16 
& 4.36 & 0.96 & \textbf{0.05}  
& 5.17 & 0.31 & -0.93 
& 4.36 & 0.20 & 0.05 \\

MetaSR \cite{hu2019meta}   
& 909.97  & 0.95  & -25.70 
& 3.93  & \underline{0.96} & \textbf{0.02}  
& 3.65 & \underline{0.98} & \textbf{-0.05} 
& 3.99 & 0.64 & -0.17 
& 4.01 & 0.44 & 0.05 \\

% MetaSR(reproduced) [48]
% & 822.32 & 0.95 & -2.43
% & 3.59 & 0.96 & 0.04
% & 3.25 & 0.98 & 0.27
% & 4.04 & 0.65 & 0.08
% & 4.09 & 0.45 & -0.02 \\

LIIF \cite{chen2021learning}     
& 808.27  & \underline{0.96}  & \underline{21.65} 
& 3.51  & \underline{0.96} & \underline{-0.07} 
& 2.97 & \underline{0.98} & 0.26  
& 3.97 & 0.65 & 0.08  
& 4.00 & 0.44 & -0.04 \\

% LIIF(reproduced)     
% & 810.75 & 0.95 & -28.16  
% & 3.53 & 0.96 & -0.25 
% & 2.94 & 0.98 & -0.38  
% & 3.99 & 0.66 & 0.08 
% & 4.05 & 0.46 & 0.02 \\

MINet \cite{chen2025arbitrary}         
& \underline{788.19}  & \underline{0.96}  & \textbf{2.28}  
& \underline{3.47}  & \textbf{0.97} & -0.10 
& \underline{2.90} & \underline{0.98} & \underline{0.20}  
& \underline{3.90} & \underline{0.66} & \underline{-0.04} 
& \underline{3.96} & \underline{0.46} & \textbf{0.00} \\

% MINet(reproduced)    
% & 885.38  & 0.95 & -32.41   
% & 4.17 & 0.95 & -0.14 
% & 5.16 & 0.94 & -0.10  
% & 4.66 & 0.56 & 0.00
% & 4.23 & 0.39 & 0.09 \\

GSASR \cite{chen2025generalized}
& 919.78 & 0.95 & -58.98
& 3.78 & \underline{0.96} & -0.36
& 3.12 & \underline{0.98} & -0.46 
& 4.23 & 0.62 & \textbf{-0.02} 
& 4.34 & 0.38 & \underline{-0.01} \\

GSSA-ViT (Ours)
& \textbf{658.58} & \textbf{0.98} & 83.23
& \textbf{3.20} & \textbf{0.97} & 0.26
& \textbf{2.82} & \textbf{0.99} & \textbf{0.05}
& \textbf{3.71} & \textbf{0.72} & -0.11
& \textbf{3.87} & \textbf{0.56} & 0.03 \\

\midrule
\multicolumn{16}{c}{MPI-ESM (5.625$^\circ$) to ERA5 (0.3515625$^\circ$)} \\
\midrule
Bicubic       
& 1142.00 & 0.92 & 72.91 
& 4.80 & 0.93 & 0.09  
& 4.12 & 0.97 & 0.30  
& 5.53 & 0.44 & -0.14 
& 5.58 & 0.20 & \textbf{0.00} \\

Bilinear      
& 1114.90 & 0.92 & 72.78 
& 4.65 & 0.94 & 0.09  
& 4.01 & 0.97 & 0.30  
& 5.29 & 0.45 & -0.14 
& 5.35 & 0.20 & \textbf{0.00} \\

\midrule

ResNet \cite{climax, he2016deep}
& 945.52  & 0.94 & 137.63 
& 4.17 & 0.95 & 0.15  
& 4.09 & 0.97 & 1.33  
& 4.59 & 0.48 & -0.53 
& 4.26 & 0.29 & 0.04 \\

Unet \cite{climax, ronneberger2015u}   
& 1025.34 & 0.93 & 138.86 
& 4.40 & 0.94 & -0.32 
& 4.42 & 0.96 & 0.82  
& 5.17 & 0.33 & -1.21 
& 4.40 & 0.16 & -0.17 \\

MetaSR \cite{hu2019meta}    
& 1026.29 & 0.94 & -35.00 
& 4.32 & 0.95 & \textbf{0.02}  
& 4.28 & 0.97 & -0.25 
& 4.05 & 0.62 & -0.20 
& 4.04 & 0.43 & 0.07 \\

% MetaSR(reproduced) [48]
% & 823.34 & 0.95 & -2.42
% & 3.60 & 0.96 & 0.04
% & 3.25 & 0.98 & 0.27
% & 4.04 & 0.65 & 0.08
% & 4.09 & 0.45 & -0.02 \\

LIIF \cite{chen2021learning}   
& 808.39  & \underline{0.96} & \underline{26.11}  
& 3.51 & \underline{0.96} & \underline{-0.07} 
& 2.96 & \underline{0.98} & 0.27  
& 3.97 & 0.65 & 0.06  
& 4.01 & 0.44 & 0.04 \\

% LIIF(reproduced)     
% & 811.63  & 0.95 & -28.00  
% & 3.53 & 0.96 & -0.25 
% & 2.95 & 0.98 & -0.37  
% & 3.99 & 0.66 & 0.08 
% & 4.05 & 0.46 & 0.02 \\

MINet \cite{chen2025arbitrary}   
& \underline{788.13}  & \underline{0.96} & \textbf{7.31 }  
& \underline{3.47} & \textbf{0.97} & -0.11 
& \underline{2.89} & \underline{0.98} & \underline{0.21}  
& \underline{3.90} & \underline{0.66} & \underline{-0.05} 
& \underline{3.96} & \underline{0.46} & \textbf{0.00} \\

% MINet(reproduced)    
% & 886.61  & 0.95 & -32.26   
% & 4.18 & 0.95 & -0.14 
% & 5.18 & 0.94 & -0.10  
% & 4.67 & 0.55 & 0.00
% & 4.23 & 0.39 & 0.09 \\

GSASR \cite{chen2025generalized}
& 920.24 & 0.95 & -48.44
& 3.78 & \underline{0.96} & -0.30
& 3.11 & \underline{0.98} & -0.38
& 4.23 & 0.62 & \textbf{0.01}
& 4.35 & 0.38 & \underline{0.02} \\

GSSA-ViT (Ours)
& \textbf{659.03} & \textbf{0.98} & 83.53
& \textbf{3.20} & \textbf{0.97} & 0.27
& \textbf{2.83} & \textbf{0.99} & \textbf{0.06}
& \textbf{3.71} & \textbf{0.72} & -0.10
& \textbf{3.87} & \textbf{0.56} & 0.03 \\

\bottomrule
\end{tabular}%
}
\end{table*}

\textbf{Pearson coefficient.} The Pearson correlation coefficient measures the linear relationship between the predicted field $\hat{X}$ and the reference field $X$:
\begin{equation}
P = \frac{\mathrm{Cov}(X, \hat{X})}{\sigma_X \, \sigma_{\hat{X}}},
\end{equation}
where $\mathrm{Cov}(X, \hat{X})$ denotes the covariance between $X$ and $\hat{X}$, and $\sigma_X$ and $\sigma_{\hat{X}}$ are their standard deviations, respectively. The coefficient ranges from $-1$ (perfect negative correlation) to $1$ (perfect positive correlation), with higher values indicating better agreement in spatial patterns.

\textbf{Mean bias.} The mean bias quantifies systematic over- or underestimation:
\begin{equation}
\text{M-b} = \frac{1}{N} \sum_{i=1}^{N} (\hat{X}_i - X_i),
\end{equation}
where $N$ denotes the total number of spatial points, positive values indicate overprediction, and negative values indicate underprediction.

\subsection{Implementation Details}
\textbf{Training Details.}
The GSSA-ViT is trained on 8 NVIDIA H200 GPUs using a data-parallel configuration. The training process consists of 200k iterations, employing the AdamW optimizer with an initial learning rate of \(1 \times 10^{-4}\). The learning rate is decayed using a cosine schedule to \(1 \times 10^{-6}\). For the arbitrary-resolution forecasting task, these 200k iterations correspond to training on 6-hourly predictions. The model is then fine-tuned with a learning rate of \(1 \times 10^{-6}\) for 36k iterations to perform 12-step (72-hour) forecasts.

% \paragraph{Evaluation Setup.}  
% The model's performance is evaluated on nine key atmospheric variables: 2-meter temperature (T2m), 10-meter zonal wind (U10), 10-meter meridional wind (V10), mean sea level pressure (MSLP), geopotential at 500 hPa (Z500), temperature at 850 hPa (T850), specific humidity at 700 hPa (Q700), wind speed (\(\sqrt{U850^2+V850^2}\)) at 850 hPa. Forecast accuracy is measured using the latitude-weighted root-mean-square error (RMSE) for lead times ranging from 1 to 14 days. GaussianCast is pretrained with a 6-hour interval, and to achieve long-term predictions, autoregressive prediction is employed for forecasts from 1 to 7 days. 

\begin{figure*}[t]
	\centering
	\includegraphics[width=\linewidth]{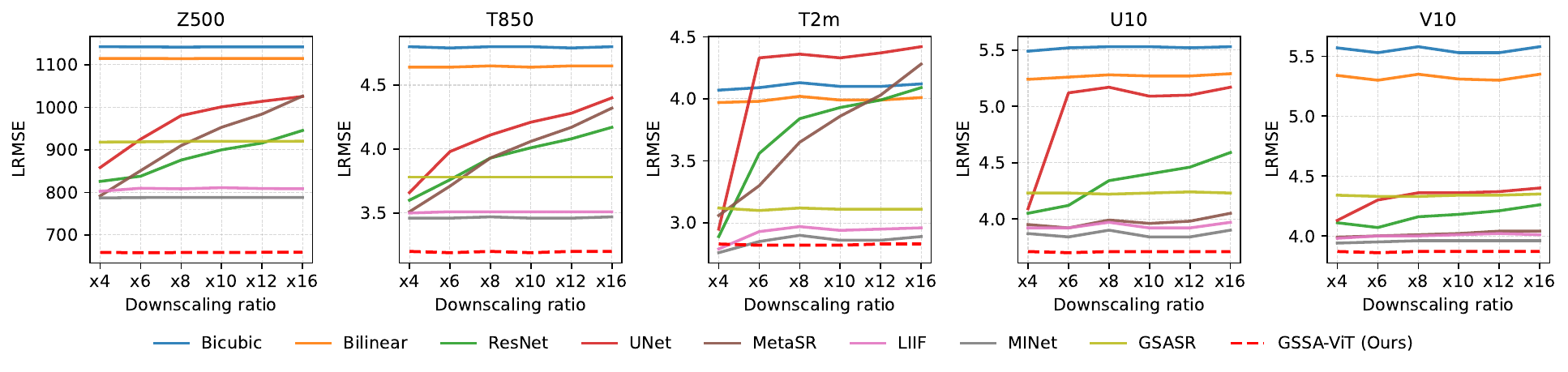}
	\caption{Performance comparison of different downscaling methods under varying downscaling ratios. The LRMSE is reported for five atmospheric variables (Z500, T850, T2m, U10, and V10) when downscaling from the CMIP (5.625°) to ERA5 targets with ratios ranging from ×4 to ×16. Lower values indicate better reconstruction accuracy.}
        \label{fig:downscaling_curve}
\end{figure*}

\begin{figure*}[!t]
	\centering
	\includegraphics[width=\linewidth]{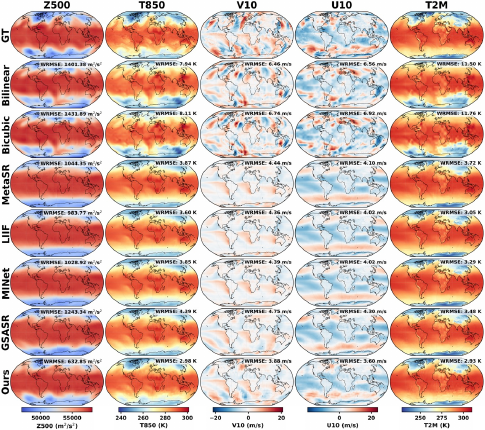}
	\caption{Global visualization of downscaling results from CMIP6 (5.625°) to ERA5 at 1.40625° resolution (×4). Each column corresponds to an atmospheric variable Z500, T850, V10, U10, and T2M. Each row shows the ground truth (GT) followed by outputs from six baselines Bilinear, Bicubic, MetaSR, LIIF, MINet, GSASR, and GSSA-ViT (Ours).}
        \label{fig:downscale4_global}
\end{figure*}

\begin{figure*}[!t]
	\centering
	\includegraphics[width=\linewidth]{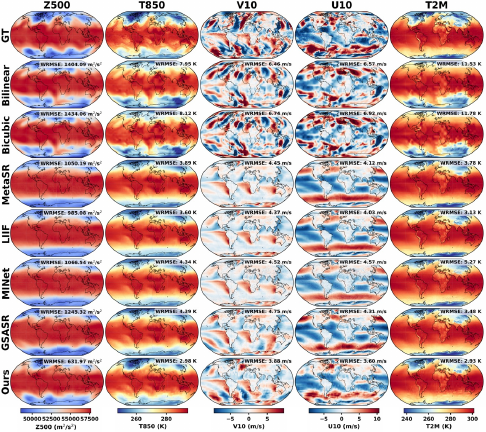}
	\caption{Global visualization of downscaling results from CMIP6 (5.625°) to ERA5 at 0.703125° resolution (×8). Each column corresponds to an atmospheric variable Z500, T850, V10, U10, and T2M. Each row shows the ground truth (GT) followed by outputs from six baselines Bilinear, Bicubic, MetaSR, LIIF, MINet, GSASR, and GSSA-ViT (Ours).}
        \label{fig:downscale8_global}
\end{figure*}

\begin{figure*}[!t]
	\centering
	\includegraphics[width=\linewidth]{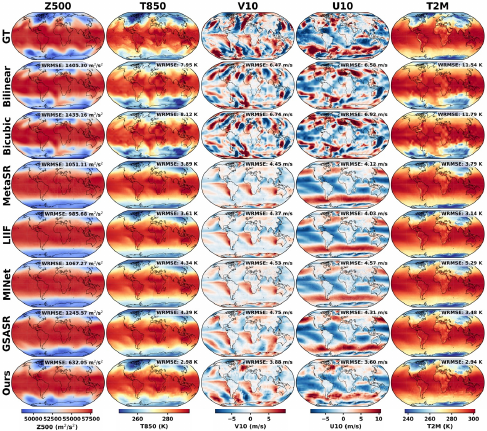}
	\caption{Global visualization of downscaling results from CMIP6 (5.625°) to ERA5 at 0.3515625° resolution (×16). Each column corresponds to an atmospheric variable Z500, T850, V10, U10, and T2M. Each row shows the ground truth (GT) followed by outputs from six baselines Bilinear, Bicubic, MetaSR, LIIF, MINet, GSASR, and GSSA-ViT (Ours).}
        \label{fig:downscale16_global}
\end{figure*}

\begin{figure*}[!t]
	\centering
	\includegraphics[width=\linewidth]{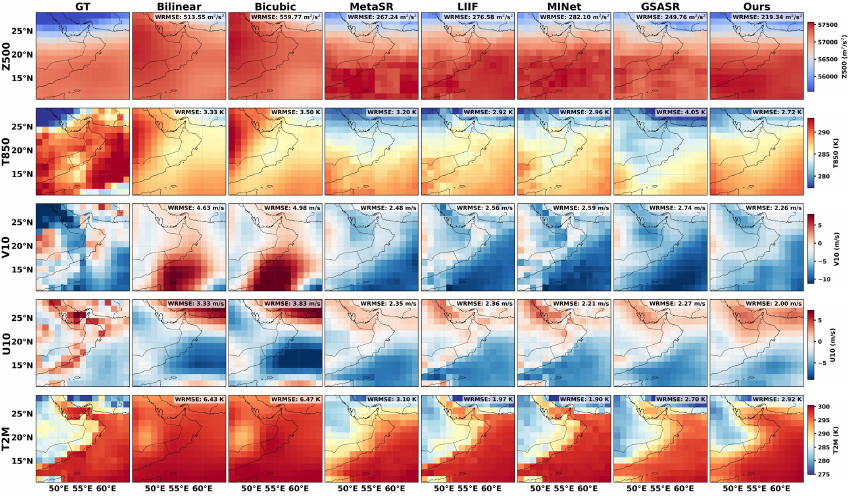}
	\caption{Regional visualization of downscaling results from CMIP6 (5.625°) to ERA5 at 1.40625° resolution (×4), focusing on the region spanning 10°–30°N and 45°–65°E. Each row corresponds to an atmospheric variable Z500, T850, V10, U10, and T2M. Each column shows the ground truth (GT) followed by outputs from six baselines Bilinear, Bicubic, MetaSR, LIIF, MINet, GSASR, and GSSA-ViT (Ours).}
        \label{fig:downscale4_local}
\end{figure*}

\begin{table*}[t]
\centering
\captionsetup{
    format=plain,           
    labelsep=newline,       
    justification=justified, 
    singlelinecheck=false,
    width=0.85\linewidth,
    labelfont=bf,
}
\caption{Performance comparison of atmospheric downscaling from ERA5 (5.625°) to ERA5 (2.8125°). Performance is evaluated using latitude-weighted RMSE (LRMSE) and mean bias (M-b) for three variables (Z500, T850, T2m). Lower LRMSE indicates better performance, while M-b values closer to zero indicate smaller bias.}
\label{tab:downscale2}
\setlength{\tabcolsep}{8pt}
\begin{tabular}{lcc cc cc}
\toprule
Methods

& \multicolumn{2}{c}{Z500} 
& \multicolumn{2}{c}{T850} 
& \multicolumn{2}{c}{T2M} \\

\cmidrule(lr){2-3}
\cmidrule(lr){4-5}
\cmidrule(lr){6-7}

& LRMSE$\downarrow$ & M-b
& LRMSE$\downarrow$ & M-b
& LRMSE$\downarrow$ & M-b \\

\midrule

Bicubic
& 269.67 & \textbf{0.04}
& 1.99 & \textbf{0.00}
& 3.11 & \textbf{0.00} \\

Bilinear
& 134.07 & \textbf{0.04}
& 1.50 & \textbf{0.00}
& 2.46 & \textbf{0.00} \\

\midrule

GSASR \cite{chen2025generalized}
& 134.44 & -76.79
& 1.23 & -0.44
& 1.79 & -0.77 \\

Unet \cite{climax, ronneberger2015u}
& 43.84 & -6.55
& 0.94 & -0.06
& 1.10 & -0.12 \\

ViT \cite{climax, vit}
& 85.32 & -35.98
& 1.03 & \underline{-0.01}
& 1.25 & -0.20 \\

LIIF \cite{chen2021learning}
& 53.79 & -3.09
& 0.96 & 0.06
& 1.07 & -0.12 \\

MINet \cite{chen2025arbitrary}
& \underline{43.61} & 1.54
& \underline{0.90} & 0.02
& \underline{0.92} & \underline{0.06} \\

GSSA-ViT (Ours)
& \textbf{41.51} & \underline{-0.06}
& \textbf{0.81} & \underline{-0.01}
& \textbf{0.82} & \underline{-0.06} \\

\bottomrule
\end{tabular}
\end{table*}

\begin{table*}[!t]
\centering
\captionsetup{
    format=plain,           
    labelsep=newline,       
    justification=justified, 
    singlelinecheck=false,
    labelfont=bf,
}
\caption{Medium-range forecasting performance at three resolutions. NeuralGCM (native resolution 1.40625$^\circ$) and Stormer (native resolution 1.40625$^\circ$) predictions are downscaled to different target resolutions using bilinear and bicubic interpolation, as well as three strong downscaling models (MetaSR, LIIF, and MINet), which are adapted to the forecasting setting by shifting ground-truth targets to the next time step. Results are reported as LRMSE for Z500, T850, Q700, and Wind850 at lead times of 6h, 24h, 72h, and 120h.}
\label{tab:forecast_multi_res_upper}
\setlength{\tabcolsep}{4pt}

\resizebox{\linewidth}{!}{
\begin{tabular}{lcccc|cccc|cccc|cccc}
\toprule

& \multicolumn{4}{c}{Z500 $\downarrow$}
& \multicolumn{4}{c}{T850 $\downarrow$}
& \multicolumn{4}{c}{Q700 $\downarrow$}
& \multicolumn{4}{c}{Wind850 $\downarrow$} \\

Lead Time
& 6h & 24h & 72h & 120h
& 6h & 24h & 72h & 120h
& 6h & 24h & 72h & 120h
& 6h & 24h & 72h & 120h \\

\midrule
\multicolumn{17}{c}{ERA5 (1.40625$^\circ$) to ERA5 (0.703125$^\circ$)} \\
\midrule
MetaSR \cite{hu2019meta}
& \underline{58.09} & 195.66 & 722.96 & 1285.11
& \underline{0.67} & 1.33 & 3.56 & 5.57
& - & - & - & -
& - & - & - & - \\

LIIF \cite{chen2021learning}
& 71.88 & 243.37 & 791.08 & 1276.04
& 0.76 & 1.55 & 3.74 & 5.58
& - & - & - & -
& - & - & - & - \\

MINet \cite{chen2025arbitrary}
& 70.64 & 231.89 & 752.94 & 1238.48
& 0.72 & 1.43 & 3.65 & 5.44
& - & - & - & -
& - & - & - & - \\

NeuralGCM \cite{Kochkov2023} (Bicubic)
& 81.92 & 95.21 & 174.56 & 333.48
& 0.86 & 1.05 & 1.41 & \textbf{1.76}
& 0.69 & 0.83 & 0.97 & 1.25
& 2.11 & 2.52 & 3.76 & 5.15  \\

NeuralGCM \cite{Kochkov2023} (Bilinear)
& 80.73 & 94.89 & 172.32 & 332.15
& 0.87 & 1.01 & 1.35 & 1.97
& 0.68 & 0.79 & 1.07 & 1.24
& 2.11 & 2.50 & 3.57 & 5.09 \\ 

Stormer \cite{nguyen2024scaling} (Bicubic)
& 78.14 & 91.80 & 170.73 & 330.23
& 0.78 & 0.93 & 1.29 & 1.88
& 0.59 & 0.72 & 0.96 & 1.17
& 2.03 & 2.40 & 3.65 & 5.01 \\

Stormer \cite{nguyen2024scaling} (Bilinear)
& 76.90 & \underline{90.62} & \underline{169.18} & \underline{328.83}
& 0.76 & \underline{0.89} & \underline{1.24} & \underline{1.85}
& \underline{0.57} & \underline{0.70} & \underline{0.93} & \underline{1.14}
& \underline{1.99} & \underline{2.36} & \underline{3.42} & \underline{4.96}  \\

GSSA-ViT (Ours)
& \textbf{39.48} & \textbf{75.94} & \textbf{158.68} & \textbf{310.71}
& \textbf{0.59} & \textbf{0.72} & \textbf{1.16} & \textbf{1.76}
& \textbf{0.46} & \textbf{0.65} & \textbf{0.86} & \textbf{1.05}
& \textbf{1.52} & \textbf{2.27} & \textbf{3.22} & \textbf{4.84} \\

\midrule
\multicolumn{17}{c}{ERA5 (1.40625$^\circ$) to ERA5 (0.3515625$^\circ$)} \\
\midrule
MetaSR \cite{hu2019meta}
& \underline{55.99} & 190.21 & 622.85 & 1029.47
& \underline{0.67} & 1.25 & 3.03 & 4.60
& - & - & - & -
& - & - & - & - \\

LIIF \cite{chen2021learning}
& 72.02 & 234.73 & 737.66 & 1161.05
& 0.76 & 1.44 & 3.28 & 4.85
& - & - & - & -
& - & - & - & - \\

MINet \cite{chen2025arbitrary}
& 72.27 & 232.15 & 687.15 & 1065.60
& 0.76 & 1.39 & 3.15 & 4.48
& - & - & - & -
& - & - & - & - \\

NeuralGCM \cite{Kochkov2023} (Bicubic)
&81.21 & 95.37 & 174.02 & 334.18
& 0.91 & 1.08 & \underline{1.19} & 1.74
& 0.66 & \underline{0.69} & 0.88 & \underline{1.11}
& 2.22 & 2.53 & 3.60 & 5.23 \\

NeuralGCM \cite{Kochkov2023} (Bilinear)
& 80.41 & 94.52 & 172.73 & 333.04
& 0.88 & 1.02 & 1.38 & \underline{1.72}
& 0.68 & 0.76 & \underline{0.83} & 1.25
& 2.13 & 2.49 & 3.55 & 5.11 \\ 

Stormer \cite{nguyen2024scaling} (Bicubic)
& 77.67 & 91.90 & 170.88 & 330.91
& 0.79 & 0.95 & 1.31 & 1.86
& 0.59 & 0.76 & 0.99 & 1.21
& 2.08 & 2.40 & 3.47 & 5.08 \\

Stormer \cite{nguyen2024scaling} (Bilinear)
& 76.88 & \underline{90.61} & \underline{169.28} & \underline{329.15}
& 0.76 & \underline{0.89} & 1.24 & 1.85
& \underline{0.56} & 0.70 & 0.93 & 1.13
& \underline{1.99} & \underline{2.36} & \underline{3.42} & \underline{4.96} \\

GSSA-ViT (Ours)
& \textbf{40.53} & \textbf{83.26} & \textbf{158.33} & \textbf{323.67}
& \textbf{0.59} & \textbf{0.84} & \textbf{1.12} & \textbf{1.69}
& \textbf{0.46} & \textbf{0.63} & \textbf{0.76} & \textbf{1.05}
& \textbf{1.54} & \textbf{2.23} & \textbf{3.28} & \textbf{4.67} \\

\midrule
\multicolumn{17}{c}{ERA5 (1.40625$^\circ$) to ERA5 (0.24965326$^\circ$)} \\
\midrule
MetaSR \cite{hu2019meta}
& \underline{62.12} & 191.55 & 617.73 & 1004.35
& \underline{0.70} & 1.26 & 2.99 & 4.50
& - & - & - & -
& - & - & - & - \\

LIIF \cite{chen2021learning}
& 72.15 & 235.68 & 752.31 & 1274.91
& 0.77 & 1.44 & 3.26 & 5.33
& - & - & - & -
& - & - & - & - \\

MINet \cite{chen2025arbitrary}
& 71.88 & 223.16 & 726.30 & 1198.12
& 0.74 & 1.38 & 3.14 & 5.09
& - & - & - & -
& - & - & - & - \\

NeuralGCM \cite{Kochkov2023} (Bicubic)
& 82.34 & 95.58 & 176.11 & 335.67
& 0.84 & 1.04 & 1.48 & 1.76
& 0.61 & 0.74 & 0.97 & 1.15
& 2.30 & 2.61 & 3.69 & 5.21  \\

NeuralGCM \cite{Kochkov2023} (Bilinear)
& 81.12 & 94.21 & 173.56 & 333.47
& 0.77 & 0.96 & 1.41 & \underline{1.74}
& 0.71 & 0.76 & \underline{0.83} & \textbf{1.10}
& 2.02 & 2.41 & 3.63 & 5.08 \\

Stormer \cite{nguyen2024scaling} (Bicubic)
& 78.87 & 91.63 & 172.20 & 332.12
& 0.90 & 0.99 & 1.34 & 1.91
& 0.62 & 0.77 & 0.96 & 1.29
& 2.18 & 2.48 & 3.55 & 5.07 \\

Stormer \cite{nguyen2024scaling} (Bilinear)
& 77.65 & \underline{90.32} & \underline{169.87} & \underline{329.85}
& 0.80 & \underline{0.91} & \underline{1.26} & 1.88
& \underline{0.58} & \underline{0.69} & 0.90 & 1.24
& \underline{1.89} & \underline{2.26} & \underline{3.48} & \underline{4.94} \\

GSSA-ViT (Ours)
& \textbf{39.57} & \textbf{79.30} & \textbf{157.98} & \textbf{321.06}
& \textbf{0.60} & \textbf{0.74} & \textbf{1.10} & \textbf{1.72}
& \textbf{0.47} & \textbf{0.58} & \textbf{0.74} & \underline{1.14}
& \textbf{1.56} & \textbf{2.13} & \textbf{3.38} & \textbf{4.75} \\

\bottomrule
\end{tabular}
}

\end{table*}

\begin{table*}[!t]
\centering
\captionsetup{
    format=plain,           
    labelsep=newline,       
    justification=justified, 
    singlelinecheck=false,  
    labelfont=bf,
}
\caption{Medium-range forecasting performance at three resolutions. NeuralGCM (native resolution 1.40625$^\circ$) and Stormer (native resolution 1.40625$^\circ$) predictions are downscaled to different target resolutions using bilinear and bicubic interpolation, as well as three strong downscaling models (MetaSR, LIIF, and MINet), which are adapted to the forecasting setting by shifting ground-truth targets to the next time step. Results are reported as LRMSE for T2M, U10, V10, and MSL at lead times of 6h, 24h, 72h, and 120h.}
\label{tab:forecast_multi_res_surface}
\setlength{\tabcolsep}{6pt}

\resizebox{\linewidth}{!}{
\begin{tabular}{lcccc|cccc|cccc|cccc}
\toprule

& \multicolumn{4}{c}{T2M $\downarrow$}
& \multicolumn{4}{c}{U10 $\downarrow$}
& \multicolumn{4}{c}{V10 $\downarrow$}
& \multicolumn{4}{c}{MSL $\downarrow$} \\

Lead Time
& 6h & 24h & 72h & 120h
& 6h & 24h & 72h & 120h
& 6h & 24h & 72h & 120h
& 6h & 24h & 72h & 120h \\

\midrule
\multicolumn{17}{c}{ERA5 (1.40625$^\circ$) to ERA5 (0.703125$^\circ$)} \\
\midrule
MetaSR \cite{hu2019meta}
& \underline{0.88} & 1.73 & 4.70 & 6.90
& \underline{0.80} & 1.65 & 3.88 & 5.12
& \underline{0.83} & 1.73 & 3.99 & 5.10
& - & - & - & - \\

LIIF \cite{chen2021learning}
& 0.98 & 2.52 & 4.41 & 5.73
& 0.90 & 1.90 & 4.52 & 6.06
& 0.94 & 1.98 & 4.67 & 6.17
& - & - & - & - \\

MINet \cite{chen2025arbitrary}
& 0.93 & 2.31 & 4.32 & 5.48
& 0.88 & 1.76 & 4.36 & 5.67
& 0.89 & 1.88 & 4.34 & 5.74
& - & - & - & - \\

Stormer \cite{nguyen2024scaling} (Bicubic)
& 1.40 & 1.42 & 1.64 & 1.91
& 1.05 & 1.26 & \underline{1.68} & 2.49
& 1.11 & 1.29 & 1.88 & \underline{2.51}
& 88.99 & 104.42 & 177.32 & 322.64 \\

Stormer \cite{nguyen2024scaling} (Bilinear)
& 1.37 & \underline{1.37} & \underline{1.58} & \underline{1.88}
& 1.01 & \underline{1.17} & 1.71 & \underline{2.48}
& 1.10 & \underline{1.25} & \underline{1.79} & 2.58
& \underline{87.62} & \underline{101.46} & \underline{176.19} & \underline{320.94} \\

GSSA-ViT (Ours)
& \textbf{0.81} & \textbf{1.00} & \textbf{1.48} & \textbf{1.78}
& \textbf{0.73} & \textbf{1.08} & \textbf{1.61} & \textbf{2.38}
& \textbf{0.74} & \textbf{1.14} & \textbf{1.68} & \textbf{2.49}
& \textbf{52.30} & \textbf{93.27} & \textbf{166.01} & \textbf{312.67} \\

\midrule
\multicolumn{17}{c}{ERA5 (1.40625$^\circ$) to ERA5 (0.3515625$^\circ$)} \\
\midrule
MetaSR \cite{hu2019meta}
& \underline{0.86} & \underline{1.38} & 3.15 & 4.99
& \underline{0.79} & 1.57 & 3.63 & 4.84
& \underline{0.83} & 1.65 & 3.77 & 4.89
& - & - & - & - \\

LIIF \cite{chen2021learning}
& 0.97 & 1.65 & 3.22 & 4.66
& 0.90 & 1.77 & 4.01 & 5.26
& 0.94 & 1.85 & 4.18 & 5.57
& - & - & - & - \\

MINet \cite{chen2025arbitrary}
& 0.99 & 1.72 & 3.65 & 5.28
& 0.91 & 1.76 & 3.86 & 5.06
& 0.95 & 1.87 & 4.07 & 5.21
& - & - & - & - \\

Stormer \cite{nguyen2024scaling} (Bicubic)
& 1.41 & 1.48 & 1.63 & 1.91
& 1.05 & 1.27 & \underline{1.68} & 2.50
& 1.11 & 1.28 & 1.88 & \underline{2.48}
& 88.97 & 104.45 & 177.32 & 322.65 \\

Stormer \cite{nguyen2024scaling} (Bilinear)
& 1.37 & \underline{1.38} & \underline{1.59} & \underline{1.89}
& 1.01 & \underline{1.18} & 1.71 & \underline{2.48}
& 1.09 & \underline{1.25} & \underline{1.79} & 2.58
& \underline{87.74} & \underline{101.63} & \underline{176.40} & \underline{321.35} \\

GSSA-ViT (Ours)
& \textbf{0.81} & \textbf{1.02} & \textbf{1.44} & \textbf{1.80}
& \textbf{0.73} & \textbf{1.04} & \textbf{1.66} & \textbf{2.33}
& \textbf{0.74} & \textbf{1.10} & \textbf{1.68} & \textbf{2.49}
& \textbf{53.39} & \textbf{95.31} & \textbf{171.04} & \textbf{320.92} \\

\midrule
\multicolumn{17}{c}{ERA5 (1.40625$^\circ$) to ERA5 (0.24965326$^\circ$)} \\
\midrule
MetaSR \cite{hu2019meta}
& \underline{0.94} & \underline{1.40} & 2.97 & 4.66
& \underline{0.82} & 1.58 & 3.61 & 4.77
& \underline{0.86} & 1.67 & 3.76 & 4.87
& - & - & - & - \\

LIIF \cite{chen2021learning}
& 1.01 & 1.58 & 2.96 & 4.98
& 0.92 & 1.78 & 4.02 & 5.99
& 0.96 & 1.86 & 4.19 & 6.44
& - & - & - & - \\

MINet \cite{chen2025arbitrary}
& 0.98 & 1.52 & 2.93 & 4.82
& 0.86 & 1.70 & 3.90 & 5.84
& 0.91 & 1.81 & 4.02 & 5.97
& - & - & - & - \\

Stormer \cite{nguyen2024scaling} (Bicubic)
& 1.43 & 1.53 & 1.76 & 1.97
& 1.09 & 1.31 & \underline{1.70} & 2.61
& 1.19 & 1.33 & 1.91 & \underline{2.52}
& 89.74 & 105.09 & 178.13 & 323.55 \\

Stormer \cite{nguyen2024scaling} (Bilinear)
& 1.39 & 1.41 & \underline{1.63} & \underline{1.94}
& 1.07 & \underline{1.20} & 1.74 & \underline{2.53}
& 1.12 & \underline{1.29} & \underline{1.84} & 2.62
& \underline{87.93} & \underline{102.78} & \underline{176.66} & \underline{321.92} \\

GSSA-ViT (Ours)
& \textbf{0.85} & \textbf{1.02} & \textbf{1.49} & \textbf{1.72}
& \textbf{0.75} & \textbf{1.14} & \textbf{1.61} & \textbf{2.39}
& \textbf{0.76} & \textbf{1.19} & \textbf{1.65} & \textbf{2.50}
& \textbf{52.62} & \textbf{92.50} & \textbf{168.09} & \textbf{319.87} \\

\bottomrule
\end{tabular}
}

\end{table*}

\textbf{Evaluation Setup.}  
For the downscaling task, we follow the evaluation protocol in \cite{chen2025arbitrary}, assessing performance on five commonly used variables: Z500, T850, T2m, V10, and U10. For methods designed for fixed-resolution inputs (e.g., ResNet and Unet), we first upsample the low-resolution inputs to the target resolution, followed by refinement using the corresponding networks.

For the arbitrary-resolution forecasting task, the model's performance is evaluated on nine key atmospheric variables: T2m, U10, V10, MSL, Z500, T850, Q700, wind speed (\(\sqrt{U850^2+V850^2}\)) at 850 hPa (Wind850). The forecasting evaluation spans lead times ranging from 1 to 3 days. GSSA-ViT is pretrained with a 6-hour interval, and to achieve long-term predictions, autoregressive prediction is employed for forecasts from 1 to 3 days. We consider two groups of baselines. First, we adapt three strong downscaling models \cite{chen2025arbitrary, hu2019meta, chen2021learning} to the forecasting setting by shifting the ground-truth targets to the next time step. Second, we include strong low-resolution forecasting baselines \cite{Kochkov2023, nguyen2024scaling}, whose outputs are upsampled to high resolution using bicubic and bilinear interpolation.

\subsection{Comparison of Arbitrary-Resolution Atmospheric Downscaling Methods}
We evaluate atmospheric downscaling from MPI-ESM (5.625°) to ERA5 at multiple target resolutions, including 1.40625°, 0.703125°, and 0.3515625°, as summarized in Table~\ref{tab:downscale1}. The corresponding performance trends across resolutions are illustrated in Fig.~\ref{fig:downscaling_curve}.

At the 1.40625° resolution, the proposed GSSA-ViT achieves the best performance across most variables and metrics. In particular, it reduces the LRMSE of Z500 to 658.84, substantially outperforming strong baselines such as MINet (786.93), MetaSR (791.71), and LIIF (802.60). Similar improvements are observed for other variables, where GSSA-ViT achieves the lowest LRMSE for T850 (3.20), U10 (3.71), and V10 (3.87), while maintaining the highest or near-highest Pearson correlations (e.g., 0.98 for Z500 and 0.99 for T2m). These results indicate that the proposed method provides more accurate reconstruction of both large-scale circulation patterns and near-surface dynamics.

As the target resolution becomes finer (0.703125° and 0.3515625°), GSSA-ViT consistently maintains superior performance compared with existing arbitrary-scale super-resolution models. For instance, at 0.703125°, our model achieves an LRMSE of 658.58 for Z500, significantly lower than MINet (788.19) and LIIF (808.27), while also achieving the highest correlations for all five variables. A similar trend is observed at 0.3515625°, where GSSA-ViT continues to outperform competing approaches with an LRMSE of 659.03 for Z500, compared with 788.13 for MINet and 808.39 for LIIF.

Fig.~\ref{fig:downscaling_curve} further highlights these advantages by showing the performance trends across different target resolutions. While most baseline methods exhibit noticeable performance degradation as the resolution becomes finer, GSSA-ViT maintains stable LRMSE, demonstrating strong robustness to resolution changes. This stability indicates that the proposed model effectively captures multi-scale atmospheric structures and generalizes well across different spatial scales.

Overall, these results demonstrate that GSSA-ViT not only achieves state-of-the-art downscaling accuracy but also provides robust performance across arbitrary target resolutions, highlighting the effectiveness of the proposed framework for high-fidelity atmospheric downscaling.

Fig.~\ref{fig:downscale4_global}, Fig.~\ref{fig:downscale8_global}, and Fig.~\ref{fig:downscale16_global} show visualized comparisons of global downscaling from CMIP6 (5.625°) to ERA5 at three target resolutions (1.40625°, 0.703125°, and 0.3515625°), including the ground truth (GT), six baselines, and GSSA-ViT (Ours). Under the 4× setting, simple interpolation methods such as bicubic and bilinear exhibit clear deficiencies, particularly over high-latitude regions. For instance, in the vicinity of Antarctica, substantial discrepancies from GT are observed across multiple variables, including Z500, T850, and T2M. Although learning-based baselines including MetaSR, LIIF, MINet, and GSASR reduce reconstruction errors relative to interpolation, especially for near-surface variables such as T2M in the Arctic, and achieve satisfactory fidelity in low- and mid-latitude regions, their ability to recover fine-grained structures in high-latitude upper-atmosphere variables such as Z500 and T850 remains limited. In contrast, GSSA-ViT consistently produces sharper and more coherent spatial patterns in these challenging regions. Furthermore, for complex surface variables such as U10 and V10, which are inherently harder to reconstruct at high resolution due to their strong spatial variability, our method still demonstrates superior detail recovery, as evidenced by the more refined U10 structures in the Arctic.

Under the 8× and 16× settings, we remove outliers in the downscaled results to improve visual clarity. The overall trends remain consistent with those observed in the 4× case. Specifically, high-latitude regions remain more challenging than low- and mid-latitude regions across all methods. Nevertheless, GSSA-ViT demonstrates clear advantages in reconstructing upper-atmosphere variables such as Z500 and T850, particularly over Antarctica, where it produces substantially sharper and more structured patterns, while MetaSR, LIIF, and MINet tend to yield overly smooth and blurred results. For complex surface variables, our method also exhibits stronger high-resolution reconstruction capability, capturing finer spatial details compared to competing approaches. This advantage is especially evident in polar regions, where spatial variability is more pronounced.

\begin{figure*}[!t]
	\centering
	\includegraphics[width=\linewidth]{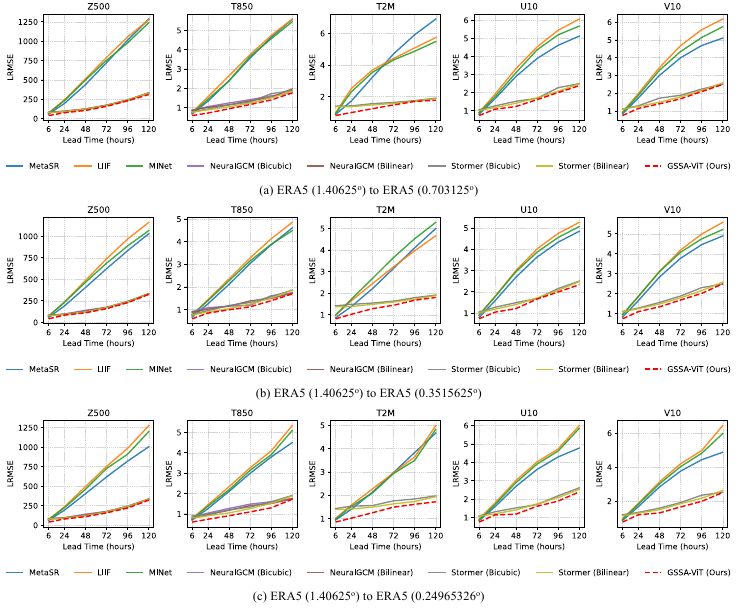}
	\caption{Performance comparison for global arbitrary-resolution prediction. The LRMSE is reported for five atmospheric variables (Z500, T850, T2M, U10, and V10). Subplots (a)–(c) correspond to predictions from ERA5 (1.40625°) to target resolutions of 0.703125°, 0.3515625°, and 0.24965326°, respectively. Results are evaluated at multiple lead times of 6, 24, 48, 72, 96, and 120 hours.}
        \label{fig:predscaling_curve_all}
\end{figure*}

\begin{figure*}[!t]
	\centering
	\includegraphics[width=\linewidth]{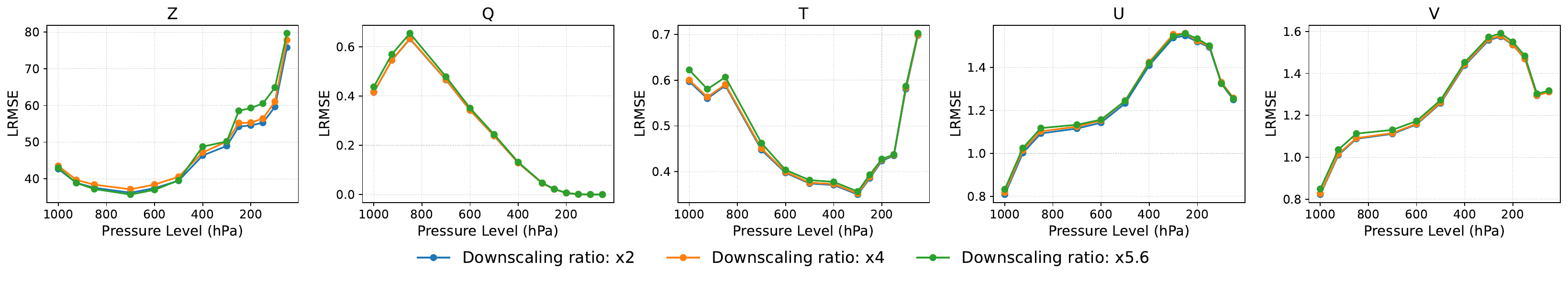}
	\caption{Performance comparison of our model under 6-hour arbitrary-resolution prediction settings across multiple atmospheric variables and vertical pressure levels. The LRMSE is reported for five variables (Z, Q, T, U, and V) as a function of pressure level (hPa), with results shown under different downscaling ratios (×2, ×4, and ×5.6). Across all variables, the model exhibits consistent performance across different downscaling ratios, with only marginal variation in error, indicating strong robustness to changes in resolution.}
        \label{fig:predscaling_curve_ours}
\end{figure*}

To provide a more detailed view at a representative scale, we further present a localized zoom-in visualization at the ×4 setting. Fig.~\ref{fig:downscale4_local} focuses on the region spanning 10°–30°N and 45°–65°E. It can be observed that interpolation-based methods produce comparatively large errors in reconstructing local structures. In contrast, our method achieves lower errors in specific localized regions and along boundaries, yielding reconstructions that are closer to the ground truth (GT) than those of existing deep learning baselines. For example, for the Z500 variable, the reconstructed field around approximately 15°N appears noticeably smoother and more consistent with the GT distribution. Similarly, for the V10 variable, the region near 15°N and 60°E shows clearer and more accurate spatial patterns, aligning more closely with the GT.

To further evaluate performance, we conduct an additional experiment by downscaling ERA5 from 5.625° to 2.8125° (×2). The quantitative results are summarized in Table~\ref{tab:downscale2}. The proposed GSSA-ViT achieves the best performance across all variables, yielding the lowest LRMSE values of 41.51, 0.81, and 0.82 for Z500, T850, and T2m, respectively. Compared with the strongest baseline MINet, our method further reduces the LRMSE from 43.61 to 41.51 for Z500, from 0.90 to 0.81 for T850, and from 0.92 to 0.82 for T2m. In addition, the mean bias remains close to zero, indicating that the proposed method not only improves reconstruction accuracy but also maintains stable statistical consistency with the reference fields. Overall, the downscaling errors from ERA5 to ERA5 are substantially lower than those from CMIP6 to ERA5 due to the differences between the datasets.

\subsection{Comparison of Arbitrary-Resolution Weather Forecasting}
The medium-range forecasting performance was evaluated on four upper-level variables: Z500, T850, Q700, and Wind850, at three target resolutions (0.703125°, 0.3515625°, and 0.24965326°), using ERA5 as the reference dataset. Latitude-weighted RMSE scores were reported for lead times of 6 hours, 24 hours, 72 hours, and 120 hours, as presented in Table~\ref{tab:forecast_multi_res_upper}. At the 0.703125° resolution, GSSA-ViT achieves 6-hour LRMSE values of 39.48 $\text{m}^2/\text{s}^2$ for Z500, 0.59 K for T850, 0.46 g/kg for Q700, and 1.52 $\text{m}/\text{s}$ for Wind850, outperforming interpolation methods and strong downscaling models including MetaSR, LIIF, MINet, NeuralGCM, and Stormer. At 24-hour and 120-hour lead times, GSSA-ViT maintains superior performance with LRMSE scores of 75.94 $\text{m}^2/\text{s}^2$ and 310.71 $\text{m}^2/\text{s}^2$ for Z500, 0.72 K and 1.76 K for T850, 0.65 $\text{g}/\text{kg}$ and 1.05 g/kg for Q700, and 2.27 $\text{m}/\text{s}$ and 4.84 $\text{m}/\text{s}$ for Wind850. At the 0.3515625° resolution, GSSA-ViT consistently achieves the lowest LRMSE values across all variables, reaching 6-hour scores of 40.53 $\text{m}^2/\text{s}^2$ for Z500, 0.59 K for T850, 0.46 $\text{g}/\text{kg}$ for Q700, and 1.54 $\text{m}/\text{s}$ for Wind850, and 120-hour scores of 323.67 $\text{m}^2/\text{s}^2$, 1.69 K, 1.05 $\text{g}/\text{kg}$, and 4.67 $\text{m}/\text{s}$. At the finest resolution, 0.24965326°, GSSA-ViT further demonstrates its advantage with 6-hour LRMSE of 39.57 $\text{m}^2/\text{s}^2$ for Z500, 0.60 K for T850, 0.47 $\text{g}/\text{kg}$ for Q700, and 1.56 $\text{m}/\text{s}$ for Wind850, and 120-hour LRMSE of 321.06 $\text{m}^2/\text{s}^2$, 1.72 K, 1.14 $\text{g}/\text{kg}$, and 4.75 $\text{m}/\text{s}$. Across all resolutions and lead times, GSSA-ViT consistently outperforms interpolation-based baselines and previous state-of-the-art downscaling models, demonstrating its effectiveness and robustness for medium-range, high-resolution atmospheric forecasting.

\begin{figure*}[!t]
	\centering
	\includegraphics[width=\linewidth]{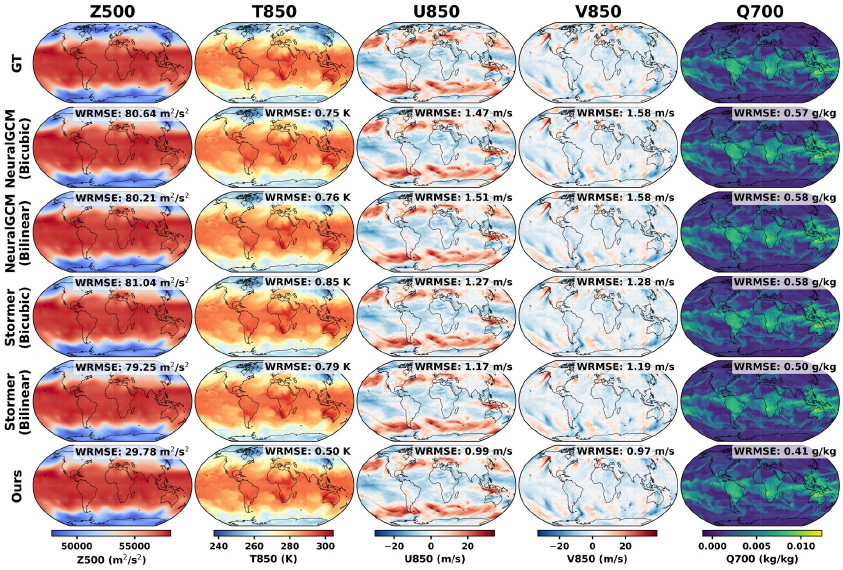}
	\caption{Global visualization of 6-hour arbitrary-resolution prediction from ERA5 at 1.40625° to 0.25° resolution. Each column corresponds to upper-level variables Z500, T850, U850, V850, and Q700. The first row shows the ground truth (ERA5 at 0.25°). Subsequent rows present results from different methods. Stormer (Bicubic) and Stormer (Bilinear) denote interpolations of Stormer predictions at native 1.40625° resolution to 0.25° using bicubic and bilinear schemes, respectively; NeuralGCM (Bicubic) and NeuralGCM (Bilinear) are defined analogously. The final row shows GSSA-ViT (Ours), which directly predicts at 0.25° resolution.}
        \label{fig:predscale5.6_global}
\end{figure*}

\begin{figure*}[!t]
	\centering
	\includegraphics[width=\linewidth]{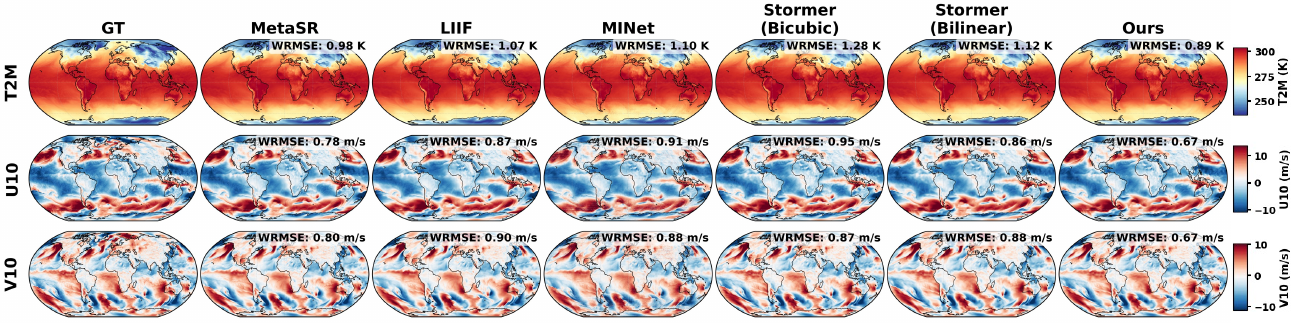}
	\caption{Global visualization of 6-hour arbitrary-resolution prediction from ERA5 at 1.40625° to 0.25° resolution. Each row corresponds to surface-level variables T2M, U10, and V10. The first column shows the ground truth (ERA5 at 0.25°). Subsequent columns present results from different methods, including MetaSR, LIIF, MINet, Stormer (Bicubic), Stormer (Bilinear), and GSSA-ViT (Ours). Stormer (Bicubic) and Stormer (Bilinear) denote interpolations of Stormer predictions at native 1.40625° resolution to 0.25° using bicubic and bilinear schemes, respectively. The final column shows GSSA-ViT (Ours), which directly predicts at 0.25° resolution.}
    \label{fig:predscale5.6_global_surface}
\end{figure*}

\begin{figure*}[!t]
	\centering
	\includegraphics[width=\linewidth]{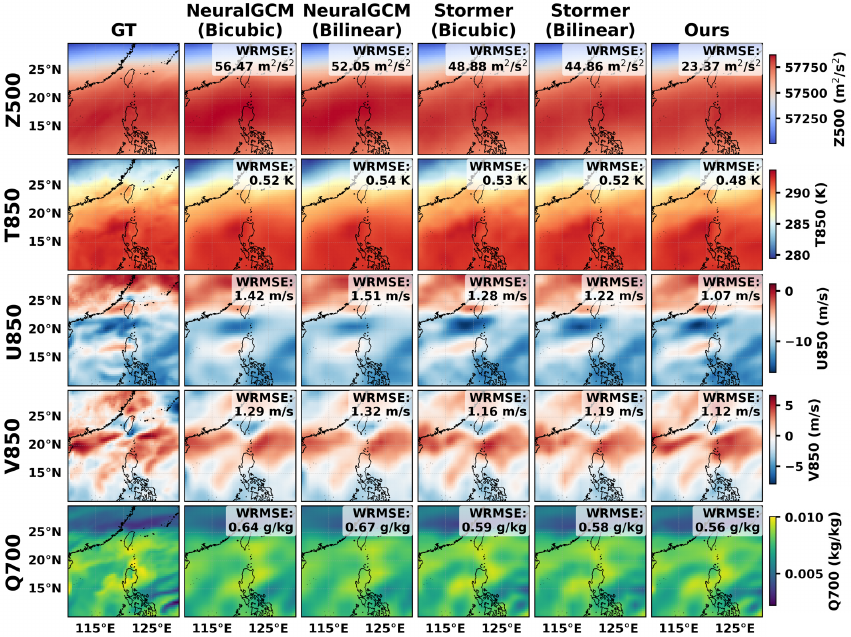}
	\caption{Regional visualization of 6-hour arbitrary-resolution prediction from ERA5 at 1.40625° to 0.3515625° resolution. Each row corresponds to surface-level variables T2M, U10, and V10. The first column shows the ground truth (ERA5 at 0.3515625°). Subsequent columns present results from different methods, including MetaSR, LIIF, MINet, Stormer (Bicubic), Stormer (Bilinear), and GSSA-ViT (Ours). Stormer (Bicubic) and Stormer (Bilinear) denote interpolations of Stormer predictions at native 1.40625° resolution to 0.3515625° using bicubic and bilinear schemes, respectively. The final column shows GSSA-ViT (Ours), which directly predicts at 0.3515625° resolution.}
        \label{fig:predscale4_local}
\end{figure*}

\begin{figure*}[!t]
	\centering
	\includegraphics[width=\linewidth]{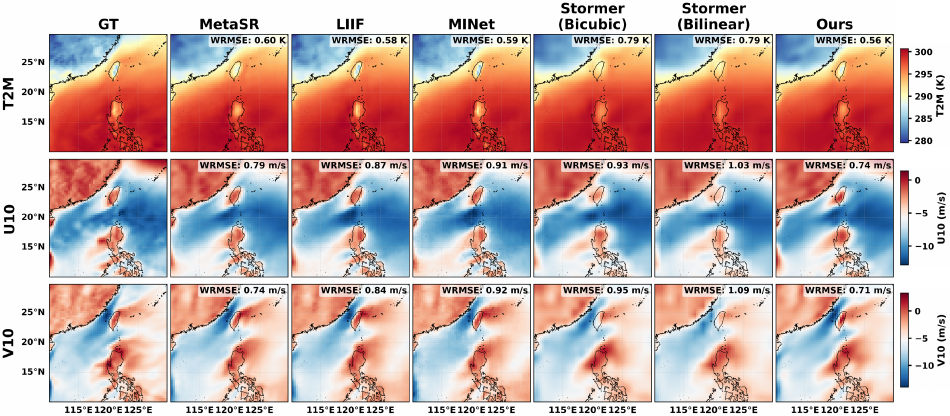}
	\caption{Regional visualization of 6-hour arbitrary-resolution prediction from ERA5 at 1.40625° to 0.3515625° resolution. Each row corresponds to surface-level variables T2M, U10, and V10. The first column shows the ground truth (ERA5 at 0.3515625°). Subsequent columns present results from different methods, including MetaSR, LIIF, MINet, Stormer (Bicubic), Stormer (Bilinear), and GSSA-ViT (Ours). Stormer (Bicubic) and Stormer (Bilinear) denote interpolations of Stormer predictions at native 1.40625° resolution to 0.3515625° using bicubic and bilinear schemes, respectively. The final column shows GSSA-ViT (Ours), which directly predicts at 0.3515625° resolution.}
        \label{fig:predscale4_local_surface}
\end{figure*}

In addition to upper-level atmospheric variables, we evaluate medium-range surface forecasting performance on four variables: T2M, U10, V10, and MSL, across three target resolutions (0.703125°, 0.3515625°, and 0.24965326°), using ERA5 as the reference dataset. Latitude-weighted RMSE scores were reported for lead times of 6 hours, 24 hours, 72 hours, and 120 hours, as shown in Table~\ref{tab:forecast_multi_res_surface}. At the 0.703125° resolution, GSSA-ViT achieves 6-hour LRMSE values of 0.81 K for T2M, 0.73 $\text{m}/\text{s}$ for U10, 0.74 $\text{m}/\text{s}$ for V10, and 52.30 Pa for MSL, significantly outperforming interpolation methods and previous strong downscaling models including MetaSR, LIIF, MINet, and Stormer. At longer lead times, GSSA-ViT maintains its advantage, reaching 120-hour LRMSE of 1.78 K for T2M, 2.38 m/s for U10, 2.49 $\text{m}/\text{s}$ for V10, and 312.67 Pa for MSL. At the 0.3515625° resolution, GSSA-ViT achieves 6-hour LRMSE values of 0.81 K, 0.73 $\text{m}/\text{s}$, 0.74 $\text{m}/\text{s}$, and 53.39 Pa, with 120-hour LRMSE scores of 1.80 K, 2.33 $\text{m}/\text{s}$, 2.49 $\text{m}/\text{s}$, and 320.92 Pa, consistently surpassing all baselines across variables and lead times. At the finest resolution, 0.24965326°, GSSA-ViT further demonstrates its effectiveness with 6-hour LRMSE of 0.85 K for T2M, 0.75 $\text{m}/\text{s}$ for U10, 0.76 $\text{m}/\text{s}$ for V10, and 52.62 Pa for MSL, and 120-hour LRMSE of 1.72 K, 2.39 $\text{m}/\text{s}$, 2.50 $\text{m}/\text{s}$, and 319.87 Pa. These results indicate that GSSA-ViT consistently outperforms both interpolation-based approaches and prior state-of-the-art downscaling models across all resolutions and forecast horizons, demonstrating its robustness and reliability for medium-range high-resolution surface weather prediction.

We further analyze the global arbitrary-resolution performance of GSSA-ViT in Fig.~\ref{fig:predscaling_curve_all}. The first set of curves presents LRMSE for five atmospheric variables—Z500, T850, T2M, U10, and V10—at target resolutions of 0.703125°, 0.3515625°, and 0.24965326° across lead times of 6, 24, 48, 72, 96, and 120 hours. GSSA-ViT consistently achieves lower errors than baseline models, with the largest improvement observed at the 6-hour lead time. As the forecast horizon increases, the performance gap gradually narrows due to the accumulation of error inherent in autoregressive prediction, yet GSSA-ViT maintains superior accuracy across all variables and resolutions, demonstrating its reliability for medium-range forecasting. Fig.~\ref{fig:predscaling_curve_ours} further evaluates the robustness of the model under different downscaling ratios (×2, ×4, and ×5.6) for multiple atmospheric variables across vertical pressure levels. The results show that GSSA-ViT maintains stable and consistent LRMSE performance across all scaling factors, with only marginal variation in error, highlighting its capability to produce accurate predictions at arbitrary resolutions while preserving consistency across both horizontal and vertical dimensions. Together, these curves confirm that GSSA-ViT not only delivers superior performance compared to existing baselines but also provides robust and scalable high-resolution forecasting across a wide range of variables, lead times, and spatial resolutions.

Fig.~\ref{fig:predscale5.6_global} and Fig.~\ref{fig:predscale5.6_global_surface} present global visualizations of arbitrary-resolution predictions from ERA5, downscaled from 1.40625° to 0.25°. For upper-level variables (Z500, T850, U850, V850, Q700) and surface-level variables (T2M, U10, V10), GSSA-ViT achieves the lowest LRMSE compared to interpolation-based baselines, NeuralGCM, Stormer, and other strong downscaling models. To examine finer spatial details, we conducted regional visualizations over the area spanning 110°–130°E and 10°–30°N, shown in Fig.~\ref{fig:predscale4_local} and Fig.~\ref{fig:predscale4_local_surface}. The results indicate that GSSA-ViT provides more accurate predictions for the V850 variable near 115°E, 20°N, and for the U10 variable near 122°E, 24°N, effectively capturing localized structures and small-scale variations. These visualizations further highlight the effectiveness of GSSA-ViT for high-resolution forecasting across both global and regional scales.

\subsection{Ablation Study}
To further verify the effectiveness of the proposed method, we conduct comprehensive ablation studies. For simplicity, all ablations are performed on the downscaling task, as we observe that the impact of each module is consistent with that in the forecasting setting. Specifically, we consider a resolution mapping from CMIP (5.625°) to ERA5 (1.40625°).

Our ablations focus on the following key aspects: (1) the function of 3D Gaussian center positioning, comparing centers fixed on latitude–longitude grid points with learnable ones; (2) the impact of Gaussian parameters, comparing fixed settings with learnable configurations for rotation, scaling, and opacity; (3) the contribution of the decoder design, comparing a unified FFN head that jointly predicts weather variables and Gaussian parameters with a two-head variant that decouples their predictions; (4) the effect of increasing the number of 3D Gaussians, implemented via an upsampling module (e.g., a lightweight convolutional layer followed by pixel shuffle) to expand the primitives to 8192; and (5) the effect of reducing the number of 3D Gaussians, achieved by using larger patch sizes in the embedding stage, decreasing the primitives to 1024. Our model uses 2048 Gaussian primitives as the default configuration. These experiments provide a concise analysis of each component’s contribution and validate our design choices.

\begin{table*}[!t]
\centering
\captionsetup{
    format=plain,           
    labelsep=newline,       
    justification=justified, 
    singlelinecheck=false,  
    labelfont=bf,
}
\caption{Ablation study on atmospheric downscaling from low-resolution CMIP data (5.6$^\circ$) to high-resolution ERA5 data (1.4$^\circ$). We report LRMSE (lower is better) across five meteorological variables, including Z500, T850, T2M, U10, and V10).}
\label{tab:ablation}

\resizebox{\linewidth}{!}{
\begin{tabular}{l c c c c | c c c c c}
\toprule
Method & Pos. Fixed & Gaussian Params Fixed & Decoder & Gaussian Num & Z500 $\downarrow$ & T850 $\downarrow$ & T2M $\downarrow$ & U10 $\downarrow$ & V10 $\downarrow$ \\
\midrule
(1) \textbf{w/o} Fixed Pos. & \ding{55} & \ding{55} & 2 heads & 2048 & 874.90 & 3.93 & 3.68 & 4.84 & 4.79\\
(2) \textbf{w/} Fixed Gaussian Params & \ding{51} & \ding{51} & 2 heads & 2048 & 930.29 & 4.05 & 3.88 & 4.76 & 4.81 \\
(3) \textbf{w/o} Gaussian Head & \ding{51} & \ding{55} & 1 head & 2048 & 678.44 & 3.32 & 2.90 & 3.82 & 3.91 \\
(4) \textbf{w/} More Gaussians & \ding{51} & \ding{55} & 2 heads & 8192 & 718.36 & 3.29 & 2.89 & 3.80 & 3.93 \\
(5) \textbf{w/} Fewer Gaussians & \ding{51} & \ding{55} & 2 heads & 1024 & 758.94 & 3.57 & 3.01  & 3.88 & 3.95 \\
GSSA-ViT & \ding{51} & \ding{55} & 2 heads & 2048 & \textbf{658.84} & \textbf{3.20} & \textbf{2.83} & \textbf{3.71} & \textbf{3.87} \\

\bottomrule
\end{tabular}
}
\end{table*}

As shown in Table~\ref{tab:ablation}, we conduct a comprehensive ablation study to evaluate the contribution of each component in the proposed framework on the downscaling task. Results indicate that fixing Gaussian center positions on latitude–longitude grids is crucial for performance. Removing this constraint leads to a substantial degradation (e.g., Z500 increases from 658.84 to 874.90, T850 from 3.20 to 3.93), suggesting that a structured spatial prior stabilizes learning and better preserves large-scale atmospheric patterns. Similarly, fixing Gaussian parameters also results in notable performance drops across all variables (e.g., Z500 increases to 930.29 and T2M increases to 3.88), demonstrating that learnable rotation, scaling, and opacity are essential for modeling complex multi-scale dynamics. In addition, adopting a two-head decoder outperforms the unified single-head variant (Z500 decreases from 678.44 to 658.84, T850 decreases from 3.32 to 3.20), indicating that decoupling weather variable prediction from Gaussian parameter estimation reduces task interference.

We further analyze the effect of the number of 3D Gaussians. Reducing the number to 1024 results in noticeable performance drops, with Z500 increasing to 758.94 and T2M increasing to 3.01, indicating insufficient representation capacity. Increasing the number to 8192 does not lead to additional improvements, as Z500 remains at 718.36 and V10 slightly increases to 3.93, which may result from increased optimization difficulty. Overall, the default configuration with 2048 Gaussians achieves the best trade-off between accuracy and efficiency, and the full model consistently outperforms all ablated variants across all variables, validating the effectiveness of each design choice.

\section{Conclusion}
In this study, we present GSSA-ViT, a unified framework for arbitrary-resolution atmospheric downscaling and medium-range forecasting based on a continuous 3D Gaussian Splatting (3DGS) representation. A key advantage of GSSA-ViT is its transition from sample-specific overfitting to a predictive, generative 3DGS paradigm, enabling accurate and computationally efficient weather prediction. This generative continuous Gaussian parameterization supports high-fidelity, localized forecasts at arbitrary spatial resolutions without requiring resolution-specific decoders or expensive physical simulations. Experimental results demonstrate that GSSA-ViT achieves state-of-the-art performance in arbitrary-resolution downscaling while maintaining highly competitive medium-range forecasting accuracy.

Despite these advantages, GSSA-ViT remains susceptible to error accumulation over extended forecast horizons, a common challenge for autoregressive AI weather models. Future work will focus on mitigating these errors by incorporating temporal consistency constraints or diffusion-based generative processes. Additionally, we plan to investigate more efficient sparse attention mechanisms to enable finer-resolution global forecasting, and explore the assimilation of ungridded operational data, such as satellite and radar observations, directly into the generative continuous Gaussian feature space. These efforts aim to enhance both the accuracy and real-world applicability of the framework, paving the way for scalable, high-fidelity weather prediction across diverse spatial and temporal scales.

\section*{Acknowledgements}
We acknowledge the founders of the ERA5 dataset and CMIP6 dataset. Without their great efforts in collecting, archiving, and disseminating the data, this study would not be possible. This work was supported by the Shanghai Artificial Intelligence Laboratory. We acknowledge the Research Support, IT, and Infrastructure team based in the Shanghai AI Laboratory for their provision of computation resources and network support. This research was supported by fundings from the Hong Kong RGC General Research Fund (152169/22E, 152228/23E, 162161/24E), Research Impact Fund (No. R5060-19, No. R5011-23), Collaborative Research Fund (No. C1042-23GF), NSFC/RGC Collaborative Research Scheme (Grant No. 62461160332 \& CRS\_HKUST602/24), Areas of Excellence Scheme (AoE/E-601/22-R), and the InnoHK (HKGAI).

% \appendix
% \section{Example Appendix Section}


%%       \cite{<label>} ==> [1]

%%
% Example citation, See \cite{lamport94}.

 \bibliographystyle{elsarticle-num} 
 \bibliography{reference}

@inproceedings{huang2024gaussianformer,
  title={Gaussianformer: Scene as gaussians for vision-based 3d semantic occupancy prediction},
  author={Huang, Yuanhui and Zheng, Wenzhao and Zhang, Yunpeng and Zhou, Jie and Lu, Jiwen},
  booktitle={European Conference on Computer Vision},
  pages={376--393},
  year={2024},
  organization={Springer}
}

@article{xu2024generalizing,
  title={Generalizing weather forecast to fine-grained temporal scales via physics-ai hybrid modeling},
  author={Xu, Wanghan and Ling, Fenghua and Han, Tao and Chen, Hao and Ouyang, Wanli and BAI, LEI},
  journal={Advances in Neural Information Processing Systems},
  volume={37},
  pages={23325--23351},
  year={2024}
}

@article{Bi2023,
  title={Accurate medium-range global weather forecasting with 3D neural networks},
  author={Bi, Kaifeng and Xie, Lingxi and Zhang, Hengheng and Chen, Xin and Gu, Xiaotao and Tian, Qi},
  journal={Nature},
  volume={619},
  number={7970},
  pages={533--538},
  year={2023},
  publisher={Nature Publishing Group UK London}
}

@article{Reichstein2019,
  title={Deep learning and process understanding for data-driven Earth system science},
  author={Reichstein, Markus and Camps-Valls, Gustau and Stevens, Bjorn and Jung, Martin and Denzler, Joachim and Carvalhais, Nuno and Prabhat, F},
  journal={Nature},
  volume={566},
  number={7743},
  pages={195--204},
  year={2019},
  publisher={Nature Publishing Group UK London}
}

@article{Brenowitz2019,
  title={Spatially extended tests of a neural network parametrization trained by coarse-graining},
  author={Brenowitz, Noah D and Bretherton, Christopher S},
  journal={Journal of Advances in Modeling Earth Systems},
  volume={11},
  number={8},
  pages={2728--2744},
  year={2019},
  publisher={Wiley Online Library}
}

@article{Kerbl2023,
  title={3d gaussian splatting for real-time radiance field rendering.},
  author={Kerbl, Bernhard and Kopanas, Georgios and Leimk{\"u}hler, Thomas and Drettakis, George},
  journal={ACM Trans. Graph.},
  volume={42},
  number={4},
  pages={139--1},
  year={2023}
}

@article{Hersbach2020,
  title={The ERA5 global reanalysis},
  author={Hersbach, Hans and Bell, Bill and Berrisford, Paul and Hirahara, Shoji and Hor{\'a}nyi, Andr{\'a}s and Mu{\~n}oz-Sabater, Joaqu{\'\i}n and Nicolas, Julien and Peubey, Carole and Radu, Raluca and Schepers, Dinand and others},
  journal={Quarterly journal of the royal meteorological society},
  volume={146},
  number={730},
  pages={1999--2049},
  year={2020},
  publisher={Wiley Online Library}
}

@inproceedings{Zhou2024,
  title={Feature 3dgs: Supercharging 3d gaussian splatting to enable distilled feature fields},
  author={Zhou, Shijie and Chang, Haoran and Jiang, Sicheng and Fan, Zhiwen and Zhu, Zehao and Xu, Dejia and Chari, Pradyumna and You, Suya and Wang, Zhangyang and Kadambi, Achuta},
  booktitle={Proceedings of the IEEE/CVF Conference on Computer Vision and Pattern Recognition},
  pages={21676--21685},
  year={2024}
}

@inproceedings{kurth2023fourcastnet,
  title={Fourcastnet: Accelerating global high-resolution weather forecasting using adaptive fourier neural operators},
  author={Kurth, Thorsten and Subramanian, Shashank and Harrington, Peter and Pathak, Jaideep and Mardani, Morteza and Hall, David and Miele, Andrea and Kashinath, Karthik and Anandkumar, Anima},
  booktitle={Proceedings of the platform for advanced scientific computing conference},
  pages={1--11},
  year={2023}
}

@article{Lam2023,
  title={Learning skillful medium-range global weather forecasting},
  author={Lam, Remi and Sanchez-Gonzalez, Alvaro and Willson, Matthew and Wirnsberger, Peter and Fortunato, Meire and Alet, Ferran and Ravuri, Suman and Ewalds, Timo and Eaton-Rosen, Zach and Hu, Weihua and others},
  journal={Science},
  volume={382},
  number={6677},
  pages={1416--1421},
  year={2023},
  publisher={American Association for the Advancement of Science}
}

@article{Chen2023,
  title={The operational medium-range deterministic weather forecasting can be extended beyond a 10-day lead time},
  author={Chen, Kang and Han, Tao and Ling, Fenghua and Gong, Junchao and Bai, Lei and Wang, Xinyu and Luo, Jing-Jia and Fei, Ben and Zhang, Wenlong and Chen, Xi and others},
  journal={Communications Earth \& Environment},
  volume={6},
  number={1},
  pages={518},
  year={2025},
  publisher={Nature Publishing Group UK London}
}

@article{Kochkov2023,
  title={Neural general circulation models for weather and climate},
  author={Kochkov, Dmitrii and Yuval, Janni and Langmore, Ian and Norgaard, Peter and Smith, Jamie and Mooers, Griffin and Kl{\"o}wer, Milan and Lottes, James and Rasp, Stephan and D{\"u}ben, Peter and others},
  journal={Nature},
  volume={632},
  number={8027},
  pages={1060--1066},
  year={2024},
  publisher={Nature Publishing Group UK London}
}

@inproceedings{Price2024,
  title={GenCast: Diffusion-based ensemble forecasting for medium-range weather},
  author={Price, Ilan and Sanchez-Gonzalez, Alvaro and Alet, Ferran and Andersson, Tom R and El-Kadi, Andrew and Masters, Dominic and Ewalds, Timo and Stott, Jacklynn and Mohamed, Shakir and Battaglia, Peter and others},
  booktitle={105th Annual AMS Meeting 2025},
  volume={105},
  pages={449275},
  year={2025}
}

@inproceedings{Zhang2023,
  title={Towards a Self-contained Data-driven Global Weather Forecasting Framework},
  author={Xiao, Yi and Bai, Lei and Xue, Wei and Chen, Hao and Chen, Kun and Chen, Kang and Han, Tao and Ouyang, Wanli},
  booktitle={International Conference on Machine Learning},
  pages={54255--54275},
  year={2024},
  organization={PMLR}
}

@article{Chen2023a,
  title={Towards an end-to-end artificial intelligence driven global weather forecasting system},
  author={Chen, Kun and Bai, Lei and Ling, Fenghua and Ye, Peng and Chen, Tao and Chen, Kang and Han, Tao and Ouyang, Wanli},
  journal={arXiv preprint arXiv:2312.12462},
  year={2023}
}

@article{Chen2024,
  title={Fengwu-ghr: Learning the kilometer-scale medium-range global weather forecasting},
  author={Han, Tao and Guo, Song and Ling, Fenghua and Chen, Kang and Gong, Junchao and Luo, Jingjia and Gu, Junxia and Dai, Kan and Ouyang, Wanli and Bai, Lei},
  journal={arXiv preprint arXiv:2402.00059},
  year={2024}
}

@article{Chen2024a,
  title={Extremecast: Boosting extreme value prediction for global weather forecast},
  author={Xu, Wanghan and Chen, Kang and Han, Tao and Chen, Hao and Ouyang, Wanli and Bai, Lei},
  journal={arXiv preprint arXiv:2402.01295},
  year={2024}
}

@article{bodnar2025foundation,
  title={A foundation model for the Earth system},
  author={Bodnar, Cristian and Bruinsma, Wessel P and Lucic, Ana and Stanley, Megan and Allen, Anna and Brandstetter, Johannes and Garvan, Patrick and Riechert, Maik and Weyn, Jonathan A and Dong, Haiyu and others},
  journal={Nature},
  volume={641},
  number={8065},
  pages={1180--1187},
  year={2025},
  publisher={Nature Publishing Group UK London}
}

@article{Lang2024,
  title={AIFS--ECMWF's data-driven forecasting system},
  author={Lang, Simon and Alexe, Mihai and Chantry, Matthew and Dramsch, Jesper and Pinault, Florian and Raoult, Baudouin and Clare, Mariana CA and Lessig, Christian and Maier-Gerber, Michael and Magnusson, Linus and others},
  journal={arXiv preprint arXiv:2406.01465},
  year={2024}
}

@article{Lang2024b,
  title={AIFS-CRPS: Ensemble forecasting using a model trained with a loss function based on the Continuous Ranked Probability Score},
  author={Lang, Simon and Alexe, Mihai and Clare, Mariana CA and Roberts, Christopher and Adewoyin, Rilwan and Bouall{\`e}gue, Zied Ben and Chantry, Matthew and Dramsch, Jesper and Dueben, Peter D and Hahner, Sara and others},
  journal={arXiv preprint arXiv:2412.15832},
  year={2024}
}

@inproceedings{climax,
  title={ClimaX: A foundation model for weather and climate},
  author={Tung Nguyen and Johannes Brandstetter and Ashish Kapoor and Jayesh K. Gupta and Aditya Grover},
  booktitle={International Conference on Machine Learning},
  year={2023},
}

@article{han2025climate,
  title={Climate science data can be compressed efficiently by dual-stage extreme compression with a variational auto-encoder transformer},
  author={Han, Tao and Chen, Zhenghao and Guo, Song and Xu, Wanghan and Ouyang, Wanli and Bai, Lei},
  journal={Communications Earth \& Environment},
  volume={6},
  number={1},
  pages={955},
  year={2025},
  publisher={Nature Publishing Group UK London}
}

@inproceedings{vit,
  title={An Image is Worth 16x16 Words: Transformers for Image Recognition at Scale},
  author={Dosovitskiy, Alexey and Beyer, Lucas and Kolesnikov, Alexander and Weissenborn, Dirk and Zhai, Xiaohua and Unterthiner, Thomas and Dehghani, Mostafa and Minderer, Matthias and Heigold, Georg and Gelly, Sylvain and others},
  booktitle={International Conference on Learning Representations},
  year={2021}
}

@article{chen2025stcast,
  title={Stcast: Adaptive boundary alignment for global and regional weather forecasting},
  author={Chen, Hao and Han, Tao and Zhang, Jie and Guo, Song and Bai, Lei},
  journal={arXiv preprint arXiv:2509.25210},
  year={2025}
}

@inproceedings{chen2025va,
  title={VA-MoE: Variables-adaptive mixture of experts for incremental weather forecasting},
  author={Chen, Hao and Tao, Han and Song, Guo and Zhang, Jie and Dong, Yonghan and Yu, Yunlong and Bai, Lei},
  booktitle={Proceedings of the IEEE/CVF International Conference on Computer Vision},
  pages={7915--7924},
  year={2025}
}

@article{mildenhall2021nerf,
  title={Nerf: Representing scenes as neural radiance fields for view synthesis},
  author={Mildenhall, Ben and Srinivasan, Pratul P and Tancik, Matthew and Barron, Jonathan T and Ramamoorthi, Ravi and Ng, Ren},
  journal={Communications of the ACM},
  volume={65},
  number={1},
  pages={99--106},
  year={2021},
  publisher={ACM New York, NY, USA}
}

@inproceedings{Cheng2024,
  title={Gaussianpro: 3d gaussian splatting with progressive propagation},
  author={Cheng, Kai and Long, Xiaoxiao and Yang, Kaizhi and Yao, Yao and Yin, Wei and Ma, Yuexin and Wang, Wenping and Chen, Xuejin},
  booktitle={Forty-first International Conference on Machine Learning},
  year={2024}
}

@inproceedings{Luiten2024,
  title={Dynamic 3d gaussians: Tracking by persistent dynamic view synthesis},
  author={Luiten, Jonathon and Kopanas, Georgios and Leibe, Bastian and Ramanan, Deva},
  booktitle={2024 International Conference on 3D Vision (3DV)},
  pages={800--809},
  year={2024},
  organization={IEEE}
}

@inproceedings{Huang2024,
  title={Sc-gs: Sparse-controlled gaussian splatting for editable dynamic scenes},
  author={Huang, Yi-Hua and Sun, Yang-Tian and Yang, Ziyi and Lyu, Xiaoyang and Cao, Yan-Pei and Qi, Xiaojuan},
  booktitle={Proceedings of the IEEE/CVF conference on computer vision and pattern recognition},
  pages={4220--4230},
  year={2024}
}

@inproceedings{Zhang2024GaussianImage,
  title={Gaussianimage: 1000 fps image representation and compression by 2d gaussian splatting},
  author={Zhang, Xinjie and Ge, Xingtong and Xu, Tongda and He, Dailan and Wang, Yan and Qin, Hongwei and Lu, Guo and Geng, Jing and Zhang, Jun},
  booktitle={European Conference on Computer Vision},
  pages={327--345},
  year={2024},
  organization={Springer}
}

@inproceedings{Zhang2024ImageGS,
  title={Image-gs: Content-adaptive image representation via 2d gaussians},
  author={Zhang, Yunxiang and Li, Bingxuan and Kuznetsov, Alexandr and Jindal, Akshay and Diolatzis, Stavros and Chen, Kenneth and Sochenov, Anton and Kaplanyan, Anton and Sun, Qi},
  booktitle={Proceedings of the Special Interest Group on Computer Graphics and Interactive Techniques Conference Conference Papers},
  pages={1--11},
  year={2025}
}

@inproceedings{vandal2017deepsd,
  title={Deepsd: Generating high resolution climate change projections through single image super-resolution},
  author={Vandal, Thomas and Kodra, Evan and Ganguly, Sangram and Michaelis, Andrew and Nemani, Ramakrishna and Ganguly, Auroop R},
  booktitle={Proceedings of the 23rd acm sigkdd international conference on knowledge discovery and data mining},
  pages={1663--1672},
  year={2017}
}

@article{mardani2025residual,
  title={Residual corrective diffusion modeling for km-scale atmospheric downscaling},
  author={Mardani, Morteza and Brenowitz, Noah and Cohen, Yair and Pathak, Jaideep and Chen, Chieh-Yu and Liu, Cheng-Chin and Vahdat, Arash and Nabian, Mohammad Amin and Ge, Tao and Subramaniam, Akshay and others},
  journal={Communications Earth \& Environment},
  volume={6},
  number={1},
  pages={124},
  year={2025},
  publisher={Nature Publishing Group UK London}
}

@article{eyring2016overview,
  title={Overview of the Coupled Model Intercomparison Project Phase 6 (CMIP6) experimental design and organization},
  author={Eyring, Veronika and Bony, Sandrine and Meehl, Gerald A and Senior, Catherine A and Stevens, Bjorn and Stouffer, Ronald J and Taylor, Karl E},
  journal={Geoscientific Model Development},
  volume={9},
  number={5},
  pages={1937--1958},
  year={2016},
  publisher={Copernicus Publications G{\"o}ttingen, Germany}
}

@article{chen2025arbitrary,
  title={Arbitrary-scale atmospheric downscaling with Mixture of Implicit Neural networks trained on fixed-scale data},
  author={Chen, Teng--Yue and Xie, Jie--Lan and Zhou, Wei and Hu, Jian--Fang and Yao, Peng--Qin and Liang, Tian-Ming and Zheng, Wei-Shi and Chan, Pak-Wai},
  journal={Pattern Recognition},
  pages={112802},
  year={2025},
  publisher={Elsevier}
}

@article{wang2026taylor,
  title={A Taylor expansion-based texture and edge-preserving interpolation approach for arbitrary-scale image super-resolution},
  author={Wang, Shibo and Xing, Yuming and Shi, Shengzhu and Guo, Zhichang},
  journal={Pattern Recognition},
  volume={169},
  pages={111965},
  year={2026},
  publisher={Elsevier}
}

@article{zhu2025multi,
  title={Multi-scale implicit transformer with re-parameterization for arbitrary-scale super-resolution},
  author={Zhu, Jinchen and Zhang, Mingjian and Zheng, Ling and Weng, Shizhuang},
  journal={Pattern Recognition},
  volume={162},
  pages={111327},
  year={2025},
  publisher={Elsevier}
}

@article{xu2019dynamical,
  title={Dynamical downscaling of regional climate: A review of methods and limitations},
  author={Xu, Zhongfeng and Han, Ying and Yang, Zongliang},
  journal={Science China Earth Sciences},
  volume={62},
  number={2},
  pages={365--375},
  year={2019},
  publisher={Springer}
}

@article{wilby1998statistical,
  title={Statistical downscaling of general circulation model output: A comparison of methods},
  author={Wilby, Robert L and Wigley, TML and Conway, D and Jones, PD and Hewitson, BC and Main, J and Wilks, DS},
  journal={Water resources research},
  volume={34},
  number={11},
  pages={2995--3008},
  year={1998},
  publisher={Wiley Online Library}
}

@article{zhang2020comparison,
  title={Comparison of statistical and dynamic downscaling techniques in generating high-resolution temperatures in China from CMIP5 GCMs},
  author={Zhang, LEI and Xu, YinLong and Meng, ChunChun and Li, XinHua and Liu, Huan and Wang, ChangGui},
  journal={Journal of Applied Meteorology and Climatology},
  volume={59},
  number={2},
  pages={207--235},
  year={2020}
}

@article{blasone2025graph,
  title={Graph neural networks for hourly precipitation projections at the convection permitting scale with a novel hybrid imperfect framework},
  author={Blasone, Valentina and Coppola, Erika and Sanguinetti, Guido and Arora, Viplove and Di Gioia, Serafina and Bortolussi, Luca},
  journal={Environmental Data Science},
  volume={4},
  pages={e47},
  year={2025},
  publisher={Cambridge University Press}
}

@article{leinonen2020stochastic,
  title={Stochastic super-resolution for downscaling time-evolving atmospheric fields with a generative adversarial network},
  author={Leinonen, Jussi and Nerini, Daniele and Berne, Alexis},
  journal={IEEE Transactions on Geoscience and Remote Sensing},
  volume={59},
  number={9},
  pages={7211--7223},
  year={2020},
  publisher={IEEE}
}

@article{hohlein2020comparative,
  title={A comparative study of convolutional neural network models for wind field downscaling},
  author={H{\"o}hlein, Kevin and Kern, Michael and Hewson, Timothy and Westermann, R{\"u}diger},
  journal={Meteorological Applications},
  volume={27},
  number={6},
  pages={e1961},
  year={2020},
  publisher={Wiley Online Library}
}

@inproceedings{liu2020climate,
  title={Climate downscaling using YNet: A deep convolutional network with skip connections and fusion},
  author={Liu, Yumin and Ganguly, Auroop R and Dy, Jennifer},
  booktitle={Proceedings of the 26th ACM SIGKDD International Conference on Knowledge Discovery \& Data Mining},
  pages={3145--3153},
  year={2020}
}

@article{wang2021deep,
  title={Deep learning for daily precipitation and temperature downscaling},
  author={Wang, Fang and Tian, Di and Lowe, Lisa and Kalin, Latif and Lehrter, John},
  journal={Water Resources Research},
  volume={57},
  number={4},
  pages={e2020WR029308},
  year={2021},
  publisher={Wiley Online Library}
}

@inproceedings{tu2025satellite,
  title={Satellite observations guided diffusion model for accurate meteorological states at arbitrary resolution},
  author={Tu, Siwei and Fei, Ben and Yang, Weidong and Ling, Fenghua and Chen, Hao and Liu, Zili and Chen, Kun and Fan, Hang and Ouyang, Wanli and Bai, Lei},
  booktitle={Proceedings of the Computer Vision and Pattern Recognition Conference},
  pages={28071--28080},
  year={2025}
}

@inproceedings{dong2016accelerating,
  title={Accelerating the super-resolution convolutional neural network},
  author={Dong, Chao and Loy, Chen Change and Tang, Xiaoou},
  booktitle={European conference on computer vision},
  pages={391--407},
  year={2016},
  organization={Springer}
}

@inproceedings{shi2016real,
  title={Real-time single image and video super-resolution using an efficient sub-pixel convolutional neural network},
  author={Shi, Wenzhe and Caballero, Jose and Husz{\'a}r, Ferenc and Totz, Johannes and Aitken, Andrew P and Bishop, Rob and Rueckert, Daniel and Wang, Zehan},
  booktitle={Proceedings of the IEEE conference on computer vision and pattern recognition},
  pages={1874--1883},
  year={2016}
}

@inproceedings{hu2019meta,
  title={Meta-SR: A magnification-arbitrary network for super-resolution},
  author={Hu, Xuecai and Mu, Haoyuan and Zhang, Xiangyu and Wang, Zilei and Tan, Tieniu and Sun, Jian},
  booktitle={Proceedings of the IEEE/CVF conference on computer vision and pattern recognition},
  pages={1575--1584},
  year={2019}
}

@inproceedings{chen2021learning,
  title={Learning continuous image representation with local implicit image function},
  author={Chen, Yinbo and Liu, Sifei and Wang, Xiaolong},
  booktitle={Proceedings of the IEEE/CVF conference on computer vision and pattern recognition},
  pages={8628--8638},
  year={2021}
}

@inproceedings{chen2025generalized,
  title={Generalized and efficient 2d gaussian splatting for arbitrary-scale super-resolution},
  author={Chen, Du and Chen, Liyi and Zhang, Zhengqiang and Zhang, Lei},
  booktitle={Proceedings of the IEEE/CVF International Conference on Computer Vision},
  pages={26435--26445},
  year={2025}
}

@inproceedings{he2016deep,
  title={Deep residual learning for image recognition},
  author={He, Kaiming and Zhang, Xiangyu and Ren, Shaoqing and Sun, Jian},
  booktitle={Proceedings of the IEEE conference on computer vision and pattern recognition},
  pages={770--778},
  year={2016}
}

@inproceedings{ronneberger2015u,
  title={U-net: Convolutional networks for biomedical image segmentation},
  author={Ronneberger, Olaf and Fischer, Philipp and Brox, Thomas},
  booktitle={International Conference on Medical image computing and computer-assisted intervention},
  pages={234--241},
  year={2015},
  organization={Springer}
}

@article{nguyen2024scaling,
  title={Scaling transformer neural networks for skillful and reliable medium-range weather forecasting},
  author={Nguyen, Tung and Shah, Rohan and Bansal, Hritik and Arcomano, Troy and Maulik, Romit and Kotamarthi, Rao and Foster, Ian and Madireddy, Sandeep and Grover, Aditya},
  journal={Advances in Neural Information Processing Systems},
  volume={37},
  pages={68740--68771},
  year={2024}
}




% %% For numbered reference style

%   \textit{\LaTeX: a document preparation system},
%   Addison Wesley, Massachusetts,

% \end{thebibliography}
\end{document}